\documentclass[preprint,12pt]{elsarticle}

\usepackage{amssymb}

\usepackage{multirow}
\usepackage{color}
\usepackage{enumitem}
\usepackage{url}
\usepackage{xcolor,colortbl}
\usepackage{stfloats}
\usepackage{booktabs}
\setlist{nolistsep}
\usepackage{url}
\definecolor{Gray}{gray}{0.9}
\usepackage{hyperref}

\usepackage{amsmath}

\begin{document}

\begin{frontmatter}



\title{Multi-Dimensional Self Attention based Approach for Remaining Useful Life Estimation}

\author[inst1]{Zhi Lai}
\ead{laizhi727@126.com}
\author[inst1]{Mengjuan Liu\corref{cor1}}
\ead{mjliu@ustc.edu.cn}
\cortext[cor1]{Mengjuan Liu is the corresponding author.}
\author[inst1]{Yunzhu Pan}
\ead{2294523327@qq.com}
\author[inst1]{Dajiang Chen}
\ead{djchen@uestc.edu.cn}

\address[inst1]{
    School of Information and Software Engineering, University of Electronic Science and Technology of China,
    Jianshe North Road, 
    Chengdu,
    610057, 
    Sichuan,
    China
}

\begin{abstract}
Remaining Useful Life (RUL) estimation plays a critical role in Prognostics and Health Management (PHM). Traditional machine health maintenance systems are often costly, requiring sufficient prior expertise, and are difficult to fit into highly complex and changing industrial scenarios. With the widespread deployment of sensors on industrial equipment, building the Industrial Internet of Things (IIoT) to interconnect these devices has become an inexorable trend in the development of the digital factory. IIoT, combined with data science, provides efficient solutions for the predictive maintenance of complex machines with multiple sensors, and the RUL prediction algorithm is the core component of these solutions. Using the device's real-time operational data collected by IIoT to get the estimated RUL through the RUL prediction algorithm, the PHM system can develop proactive maintenance measures for the device, thus, reducing maintenance costs and decreasing failure times during operation. This paper carries out research into the remaining useful life prediction model for multi-sensor devices in the IIoT scenario. We investigated the mainstream RUL prediction models and summarized the basic steps of RUL prediction modeling in this scenario. On this basis, a data-driven approach for RUL estimation is proposed in this paper. It employs a Multi-Head Attention Mechanism to fuse the multi-dimensional time-series data output from multiple sensors, in which the attention on features is used to capture the interactions between features and attention on sequences is used to learn the weights of time steps. Then, the Long Short-Term Memory Network is applied to learn the features of time series. We evaluate the proposed model on two benchmark datasets (C-MAPSS and PHM08), and the results demonstrate that it outperforms the state-of-art models. Moreover, through the interpretability of the multi-head attention mechanism, the proposed model can provide a preliminary explanation of engine degradation. Therefore, this approach is promising for predictive maintenance in IIoT scenarios.\par

\end{abstract}



\end{frontmatter}

\section{Introduction}
The advent of low-cost micro-sensors and high-bandwidth wireless networks allows us to access machine status data in real-time across physical space. Meanwhile, big data and artificial intelligence technology can mine valuable information from massive data to improve productivity and reduce the risk of failure. Driven by these rising technologies, the Industrial Internet of Things (IIoT) emerged as the times require\cite{iiotai5g}, transforming data from a by-product of the manufacturing process to a strategic resource of great concern to companies. One highly promising application in IIoT is to apply big data technology to the predictive maintenance of complex mechanical equipment\cite{iiotphm}.\par

Mechanical failures do not occur regularly for complex mechanical equipment, resulting in higher maintenance time and cost, and these failures usually cause catastrophic consequences, such as aircraft engine damage\cite{survey06}. Predictive maintenance monitors equipment's operating parameters through sensors embedded in it, uses big data to analyze the cycle, area, and failure types, which is used to predict equipment's RUL, thus, can repair it in advance, ensuring the stability and safety of the system. So far, predictive maintenance has been widely used in many fields such as aerospace, naval, power, gradually replacing event-driven breakdown maintenance and time-driven scheduled maintenance.\par

RUL estimation algorithm is the core algorithms for predictive maintenance and PHM systems. An accurate RUL prediction can significantly reduce repair and maintenance costs while also minimize the risk of unplanned shutdowns\cite{survey11}. Moreover, RUL estimation also plays an essential role in other areas, such as product reuse and recycling management, energy consumption, and landfills\cite{survey07}. As a result, the significance of RUL estimation has gone beyond PHM to become a key algorithm in several industrial fields. RUL estimation algorithms can be categorized into physics model-based, data-driven, and hybrid algorithms. The physical model-based approach builds a precise model with physical laws, while the data-driven approach makes estimation by analyzing the data\cite{pecht2009survey}. Data-driven algorithms can be further categorized into statistical-based methods and machine-learning-based methods. Statistical approaches include Markov models, Weibull distribution, Kalman filters, etc. In addition, machine learning approaches, including logistic and regression neural networks, train the models to convergence by gradient descent. Finally, the hybrid approach introduces physical modeling into data-driven methods, achieving better performance in scenarios where expertise is accessible\cite{pecht2010survey}.\par

In recent years, data-driven RUL prediction algorithms have received increasing attention. Due to the modern mechanical systems becoming more and more complex, it is nearly impossible to build accurate physical prediction models based on failure mechanisms. Instead, data-driven algorithms generally use an end-to-end design that enables to learn trends in machine degradation directly from sensor data and obtain results directly without concern for intermediate processes\cite{nn08}. Moreover, data-driven approaches do not require prior expertise in specific fields, thus, are more conducive to industrial deployment. Also, data-driven deep learning approaches are more suitable for massive amounts of data in IIoT scenarios. As the amount increases, deep learning algorithms yield more accurate results than traditional machine learning methods\cite{iiotrul1}\cite{iiottcn}.\par

Traditional machine learning algorithms (such as SVM\cite{svr}, ANN\cite{nn09}, and DBN\cite{dbn}) have achieved good results in RUL prediction. However, these methods are gradually displaced by neural networks because they are challenging to extract abstract features and do not scale well in big data scenarios. Convolutional neural networks (CNNs) have been proved to be efficient in extracting features from 1D and 2D data and are widely used in the industry for fault detection and RUL prediction\cite{cnn1d}. Another prevailing structure is the recurrent neural networks (RNNs)\cite{rnn08}. In the RUL prediction task, the data are generally composed of time series output from multiple sensors. RNNs do well in extracting feature information of the time series, which is not available in CNNs. \par

However, these networks all treat the time series of multiple sensors equally, but in fact, different sensors have different contributions to the RUL values. The attention mechanism was used in machine translation tasks\cite{bahdanau2016neural}\cite{kim2017structured}, showing a good performance. Its interpretable weight settings have led to its widespread use in areas such as images\cite{hu2019squeezeandexcitation} and recommendations\cite{din}. The attention mechanism first adds a layer of trainable parameters to the last step's network and then regularizes them into interpretable weights using the softmax function. In this way, the features of the previous step are weighted, and in the meantime, interpretability is achieved. Some models incorporating attention mechanisms have been proposed for the RUL prediction task. According to where the attention mechanisms are used, they are divided into three types: time weighting, feature (sensor) weighting, and two dimensions weighting together. The above methods all use basic attention mechanisms. Nevertheless, in RUL degradation, there may be multiple patterns of sensor degradation, and the basic attention mechanism may not be able to take these degradation paths into account simultaneously. Furthermore, the basic attention mechanism cannot model the intrinsic connection of features.\par

A deep learning method based on the multi-head self-attention mechanism is proposed in this paper to solve those problems. Unlike the traditional attention mechanism, the self-attention mechanism does not require external information and is better at capturing the internal correlations of features. Additionally, the multi-head attention mechanism uses the attention weights of multiple dimensions to aggregate different contributions of features from multiple perspectives. Therefore, it is more generalizable. The proposed method learns both feature interactions between sensors and weights between sequences. First, it utilizes a self-attention mechanism on features to learn interactions between model features and a self-attention mechanism on sequences to learn the influence weights of different time steps. Then, an LSTM network is set up for learning time-series features. To validate the effectiveness of our approach, we conducted experiments on two benchmark datasets and compared our method with some state-of-the-art models.\par

The main contributions of this paper are summarized as follows:

\begin{itemize}

\item We propose an advanced data-driven model for RUL estimation, which introduces the multi-dimensional multi-head self-attention mechanism to learn data features from various aspects. The proposed model learns interactions between features by a self-attention mechanism on features, then learns the influence weights of different time steps by another self-attention mechanism on sequences, and finally uses LSTM to process multi-dimensional sequence features.

\item We conduct extensive experiments on two widely used datasets. The experimental results demonstrate that the proposed model outperforms the state-of-art models. In addition, we present a preliminary analysis of the interpretability of the proposed model.

\item We share our experience in dealing with some practical issues in our experiments and discuss and analyze the settings and effects of some critical parameters in detail. Furthermore, we open-source the experiment code, which we believe will help the community develop.

\end{itemize}

\section{Related Works}
Early RUL prediction algorithms usually used model-based approaches. In \cite{bolander2009physics}, the RUL of the aircraft engine bearings is estimated using a bearing spall propagation model with the particle filter-based approach, which utilized vibration and online oil debris sensors to detect spalling of bearings. Coppe et al.\cite{coppe2012using} use a Paris model with an assumed analytical stress-intensity factor to estimate the remaining useful life of a system experiencing fatigue crack growth. Cheng and Pecht\cite{fusion_pecht} proposed a hybrid model for predicting the RUL of electronic products. A data-driven approach is used to detect faults and extract features, and the PoF model with a data-driven method is used to make predictions.\par

In recent years, machine learning methods have been widely applied to problems in manufacturing and industrial systems, and have achieved excellent results in RUL prediction. For example, support vector machines (SVM)\cite{svr}, artificial neural networks (ANN)\cite{nn09}\cite{nn08} and deep belief network (DBN)\cite{dbn} algorithms have been widely used for RUL prediction.\par

With the growing development of machine learning algorithms, neural networks have successfully become the mainstream, among which the first to stand out is the CNN. Sateesh Babu et al.\cite{cnn1d} introduced a deep CNN-based regression approach for the turbine engine’s RUL estimation and proved the CNN outperformed several traditional algorithms such as MLP, SVR, and RVR on two publicly available datasets. Li et al.\cite{cnn2d} also proposed a deep convolution neural network(DCNN) method. Different from the previous method using a 1D CNN, they used a 2D deep CNN to predict RUL. In RUL prediction, CNNs that use large convolutional kernels can extract abstract features directly from the raw data, and the sparse connection and weight sharing strategies allow the network to stack many layers without worrying about gradient vanishing so that the ability to extract features is improved while training parameters are reduced. Thus, CNNs achieve better results than traditional machine learning methods.\par

However, CNNs ignore the temporal characteristics of the data. In CNNs, the previous input is entirely unrelated to the following input, while most of the sensor data are time-series data, so RNN, a network designed to process time-series data, is naturally used by many researchers. RNNs can remember every time step's information, and the hidden layer at each time step is determined not only by the input layer at that moment but also by the hidden layer at the previous time step. Zheng et al.\cite{lstm1} used LSTM for the RUL prediction task, and the experimental results obtained on three datasets confirmed the better performance of LSTM over traditional methods as well as CNNs. Liao et.al\cite{lstm2} proposed an LSTM based on Bootstrap for uncertainty prediction of RUL estimation. Elsheikh et al.\cite{bilstm} replaced LSTM with bidirectional LSTM to achieve RUL prediction for random starting short sequences of monitored observations and proposed a safety-oriented objective function to train the network to favor safer early predictions rather than later predictions. Al-Dulaimi et al.\cite{hdnn} proposed a method called HDNN, which mixed LSTM and CNN networks, using LSTM to extract temporal features and CNN to extract spatial features, and the two networks were integrated in parallel.\par

Graph structure and graph neural networks develop rapidly recently and are widely used in image recognition, recommendation, and other fields. Compared with the hierarchical structure of traditional neural networks, graph structures can express more complex relationships. Some studies have also used graph neural networks in the RUL prediction task, where good results have been achieved. Li et al.\cite{dag} proposed a graph structure-based method to predict RUL. The method used directed acyclic graphs to combine LSTM with CNN networks organically instead of simply stacking or integrating LSTM and CNN. Narwariya et al.\cite{gnmr} used graph structures to capture the complex structures inside the device to improve the performance of RUL prediction. The approach used a gated graph neural network (GGNN) to model the internal module structure of the turbine engine, thus dividing the multi-dimensional time-series data into meaningful subsets.\par

Most deep learning algorithms do not consider that different input features contribute differently to the RUL values. However more important features should be paid more attention to help the model focus more on the critical data so that the results can be more accurate. The Attention mechanism is used to achieve this and has been widely applied to NLP, recommendation, and many other areas with excellent results.\par
Das et.al \cite{attn_bilstm} introduced an attention-based bidirectional LSTM network model that weights the time steps. They considered that LSTM only used the output of the last time step for RUL prediction, which might lose the information from the earlier time steps. Xia et al.\cite{mlsa_lstm} implemented the self-attention mechanism of Transformer's Encoder module to weigh the time steps of the input data. The attention layers and linear layers are alternately arranged and connected using shortcut connections. Ragab et al.\cite{ats2s} proposed an LSTM based sequence to sequence model with an Attention mechanism for RUL prediction. Cao et al.\cite{residual_attention_tcn} proposed a model named TCN-RSA for RUL estimation of rolling bearings,  which consists of a temporal convolutional network append with a residual self-attention mechanism in the time dimension. Liu and Wang\cite{deep_and_attention} proposed an RUL prediction model based on CNN and self-attention using two paralleled network structures. The left side is a DNN, and the right side is first a CNN and position encoding to process the temporal data, then the time series are weighted using the attention mechanism. Xu et al.\cite{ds_sann} applied a dual-stream self-attention model, which uses two attention networks in parallel to extract features. In \cite{attn_cnn_gru}, a feature-attention-based model (AGCNN) is presented to predict the RUL of turbofan engines. The author proposed to apply the attention mechanism directly to the input data, where the more critical features were given greater weights. Then the weighted data were passed into Bi-GRU and CNN to extract long-term dependencies and capture local features. \par
The articles referenced above only considered the importance of either time steps or features. However, different time steps and features have different RUL contributions so that weighting can extract even more critical information and more accurate results. Chen et al.\cite{attn_lstm} proposed an attention-based LSTM deep learning framework. They used an attention mechanism to weight the 2D sequence data in time dimension output by LSTM, achieving that time steps and features were weighted simultaneously.\par
Song et al.\cite{tcn} presented a distributed attention-based temporal convolutional network (TCN) for RUL prediction. The first attention mechanism passed sensor data collected simultaneously into the softmax function to calculate the weight of each feature at that moment, and the final weight of each feature is the average of the weights obtained from all time steps in the time window. Then the data are multiplied by the weights and passed into TCN for feature extraction, and the outputs are weighted by another attention mechanism in the time dimension. \cite{dual_transformer} applied the fully Transformer's structure to RUL prediction. A position encoding layer is used before the encoder to parse the timing information. Then two encoder modules process the input in parallel that process the input from both time and sensor dimension. Finally, the two Encoder outputs are connected and sent into the Decoder and then sent to the linear layer to output the result.\par

Most of the above attention mechanisms directly regularize the parameters to interpretable weights by softmax. With the research of attention mechanisms going further and further, many fundamental models of attention mechanisms have been proposed, among which multi-head attention is one of the best. Multi-head attention mechanism transforms the input linearly and then feeds it into different attention networks separately, without sharing parameters between different heads, so that data features can be learned from multiple perspectives. The self-attention mechanism is another variation of the attention mechanism. Self-attention focuses more on the interactions between features and can explore feature interactions that ordinary attention cannot capture. Our model is based on the multi-head self-attention mechanism and weights both time steps and features. Thus  we can better learn the feature interactions and degradation trends among multi-sensor data.\par

\section{Methodology}



\subsection{Multi-Head Attention} \label{section_multi_head_attention}
\begin{figure*}[htbp]
  \centering
  \includegraphics{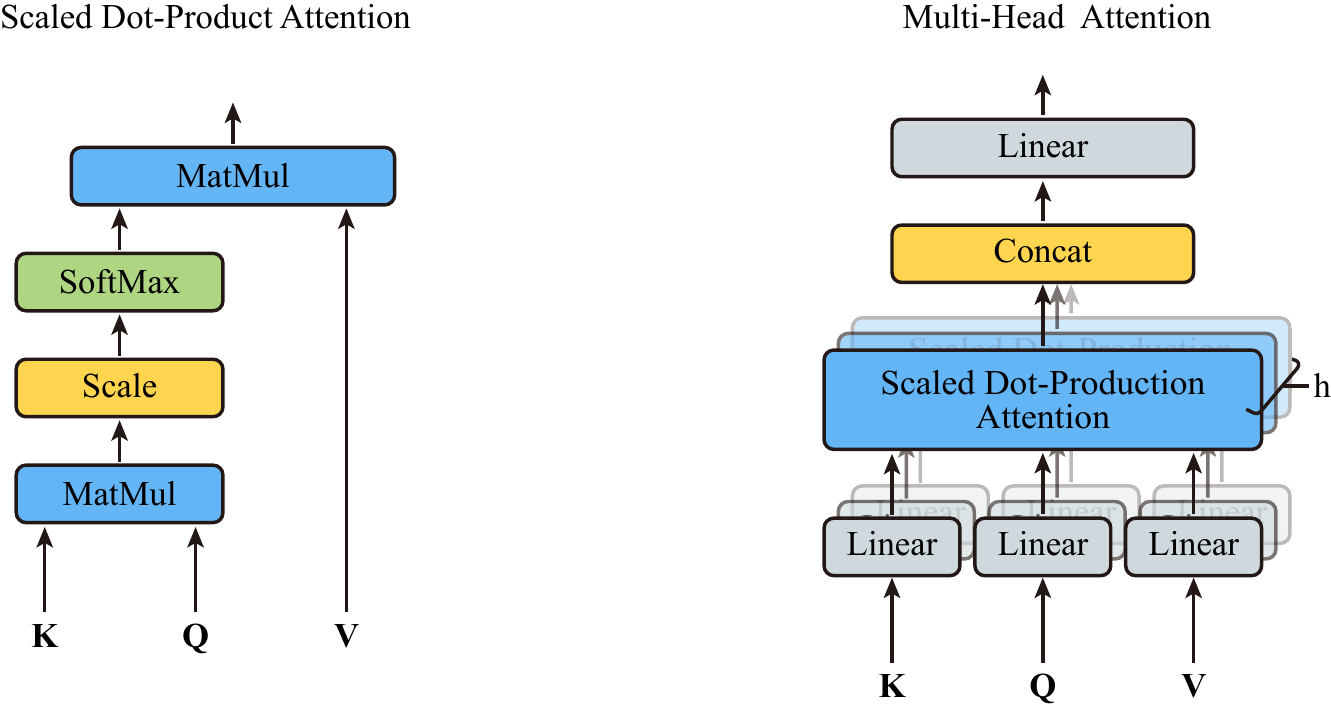}
  \caption{Scaled Dot-Product Attention (left) and Multi-Head Attention Mechanism (right) diagram\cite{multi_head_attn}.}
  \label{multi_head_attention}
\end{figure*}
The multi-head attention mechanism was proposed in \cite{multi_head_attn} and has achieved remarkable results in machine translation. The attention mechanism can be abstracted as a mapping function from a $Query$ to a series of key-value pairs. The similarity between $Query$ and $Key$ is calculated using the scaled dot product, reflecting the importance of $Value$, i.e., attention weight. The formula of scaled dot-production attention is as follows.
 \begin{equation}
Attention(Q,K,V) = softmax(\frac{QK^T}{\sqrt{d_k}})V
\end{equation}
where $d_k$ denotes the dimension of $Key$, he scaling factor $\sqrt{d_k}$ is used to solve the vanishing gradient problem when $d_k$ is large, which helps distribute attention weights and get a better generalization effect.\par

Multiple scaled dot-production attentions are first connected in multi-head attention, and then a weight matrix is used to map the output to the size of a single head. The formula is defined as
\begin{equation}
\begin{split}
MultiHead(Q,K,V) &= Concat(head_1,head_2,...,head_h)W^O\\
where\ head_i &= Attention(QW_i^Q, KW_i^K,VW_i^V)
\end{split}
\end{equation}
where $h$ denotes the number of heads, $W^Q$, $W^K$, $W^V$, and $W^O$ are all trainable parameter matrices. Each $head$ is a scaled dot product attention, but the weights obtained in training are different. So multi-head attention can be considered as ensemble learning, aggregating the results of various attention mechanisms with different perspectives. A schematic representation of the scaled dot-production attention and the multi-head attention is shown in Figure \ref{multi_head_attention}.\par

\subsection{Proposed Method}
\begin{figure*}[htbp]
  \centering
\resizebox{\columnwidth}{!}{
    \includegraphics{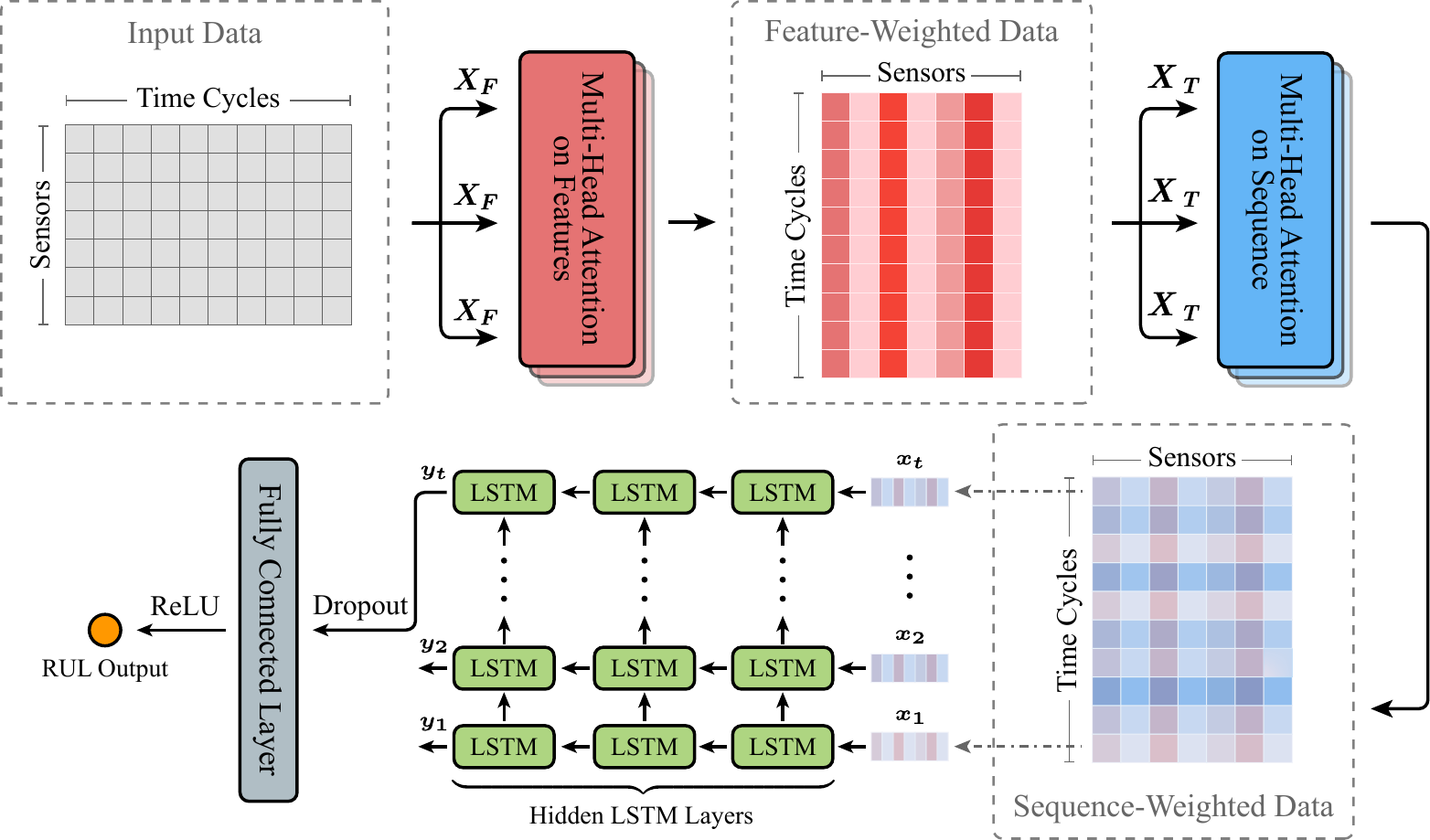}
}
  \caption{Overall structure of the proposed method.}
  \label{model_structure}
\end{figure*}

An overview of the proposed model is shown in Figure \ref{model_structure}. First, the model input is a multi-dimensional time series that is produced by multiple sensors. The inputs can be represented as an $F \times T$ matrix representing $T$-step time series from $F$ sensor inputs.\par

The input matrix with the dimension of $F \times T$ is first fed to the multi-head self-attention mechanism. The $T$ time steps' data of each sensor can be considered as the embedding of this sensor in the $T$ dimensions, which is expressed as
\begin{equation}
    X_F = [S_1,S_2,...,S_F],\ S_i = [s^i_{1}, s^i_{2},...,s^i_{T}]
\end{equation}
where $F$ is the number of sensors, $T$ is the number of time steps, $S_i$ indicates the time series of a single sensor, and $s^i_j$ represents the data of the $i$-th sensor at the $j$-th time step. $j=1,2,...,T$. In self-attention, $Q=K=V$, the self-attention in the sensor dimension can be represented as
\begin{equation}
    Output = MultiHead(X_F; X_F; X_F )
\end{equation}
where $MultiHead$ represents the multi-head attention mechanism introduced in section \ref{section_multi_head_attention}, and $X_F$ represents the input of the attention mechanism. The feature map weighted by the sensor is obtained after the attention layer.\par

The output of the attention mechanism in sensor dimension won't change its dimension and then is weighted by the attention in sequence dimension. In this step, we consider all sensor data in a time step as the embedding of this time step, so the matrix of the input can be expressed as
\begin{equation}
    X_T = Out_F = [O_1,O_2,...,O_T],\ O_j = [o^j_1, o^j_2,...,o^j_F]
\end{equation}
And the attention mechanism can be defined as
\begin{equation}
    Output_T = MultiHead(X_T; X_T; X_T)
\end{equation}
After these two steps, we get a time series weighted in both sequence and sensor dimensions.\par

The weighted data are then fed into a deep LSTM network to learn the sequence features. By stacking LSTM cells vertically, the LSTM network can learn more high-order sequence features. For each time step of LSTM network, the input of upper LSTM cells is the output of lower cells. The output of the last LSTM layer at the last time step is used as the feature vector, which has contained the information of all previous time steps.\par

The feature vector is sent through a multilayer perceptron (MLP) and mapped to individual neuron outputs to obtain the final RUL estimation. The dropout\cite{dropout} was used to mitigate the overfitting problem. In the dropout technique, neurons are randomly masked with probability $P$ during training, while during testing, all neurons are involved in the output.\par

Considering that RUL estimation is a typical regression problem, we use the mean square error as the loss function.
\begin{equation}
    L=\frac{1}{N} \sum_{i=1}^N(y_i - \hat{y_i})^2
\end{equation}
where $y_i$ and $\hat{y_i}$ denote the real and prediction values of RUL, respectively, and $N$ is the total number of train set samples.
\par

The proposed model is an end-to-end model with all parameters trained jointly. Batch gradient descent algorithms are employed to update the parameters, and the early-stop mechanism is used to prevent overfitting. The details of hyper-parameters will be presented in section \ref{section_hyper_param}.\par

\section{Experiment Setting}
\subsection{Dataset}
The turbofan engine degradation simulation dataset(C-MAPSS dataset)\cite{cmapss} is one of the most popular public datasets for RUL estimation tasks provided by NASA. The C-MAPSS dataset includes four sub-datasets that collect degradation data for aircraft engines simulated by C-MAPSS in two fault modes and six operating conditions. The PHM08 dataset comes from the challenge at the first PHM conference in 2008 and has the same format as the C-MAPSS dataset, except that the C-MAPSS dataset reveals the results of the test set, while the PHM08 dataset requires the researcher to upload the results to the official server via a webpage\footnote{\url{https://ti.arc.nasa.gov/tech/dash/groups/pcoe/prognostic-data-repository/}} to get the test results so that the results of PHM08 are more objective and fair compared to C-MAPSS.\par

\begin{figure}[htbp]
  \centering
  \includegraphics[width=0.5\textwidth]{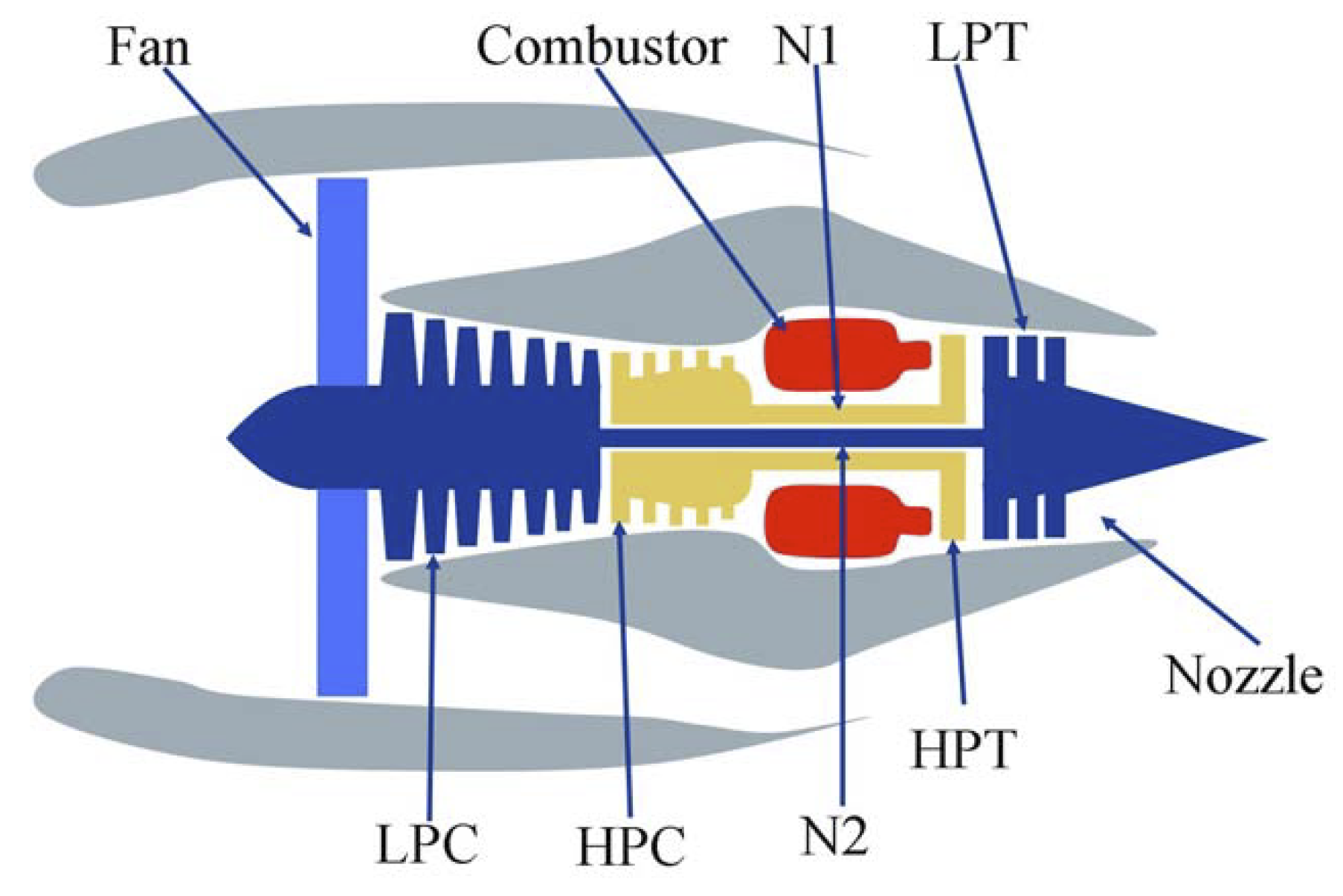}
  \caption{Simplified diagram of turbofan engine simulated in C-MAPSS\cite{cmapss_user_guide}}
    \label{turbofan_diagram}
\end{figure}

\begin{table}[htbp]
  \centering
  \caption{C-MAPSS Dataset Statistics}
  \begin{tabular}{lrrrrr}
    \toprule
    \multicolumn{1}{c}{\multirow{2}[1]{*}{Dataset}} & \multicolumn{4}{c}{C-MAPSS} &
   \multicolumn{1}{c}{\multirow{2}[1]{*}{PHM08}} \\
    \cmidrule{2-5}
          & \multicolumn{1}{l}{FD001} & \multicolumn{1}{l}{FD002} & \multicolumn{1}{l}{FD003} & \multicolumn{1}{l}{FD004}  \\
    \midrule
    Train trjectories & 100   & 260   & 100   & 248 & 218 \\
    Test trjectories & 100   & 259   & 100   & 249& 218 \\
    Conditions & 1     & 6     & 1     & 6& 6 \\
    Fault modes & 1     & 1     & 2     & 2& 1 \\
    \bottomrule
\end{tabular}%
  \label{cmapss_dataset_statistic}%
\end{table}%

\begin{table}[htbp]
  \centering
  \caption{Sensor and Environment parameters explanation in C-MAPSS Dataset\cite{cmapss}}
  
\begin{tabular}{lrr}
    \toprule
    Sensor Symbol & Description & Units \\
    \midrule
    T2 & Total temperature at fan inlet & R\\
    T24 & Total temperature at LPC outlet& R\\
    T30 & Total temperature at HPC outlet & R\\
    T50 & Total temperature at LPT outlet & R\\
    P2 & Pressure at fan inlet & psia\\
    P15 & Total pressure in bypass-duct & psia\\
    P30 & Total pressure at HPC outlet & psia\\
    Nf & Physical fan speed & rpm\\
    Nc & Physical core speed & rpm\\
    epr & Engine pressure ratio (P50/P2) & --\\
    Ps30 & Static pressure at HPC outlet & psia\\
    phi & Ratio of fuel flow to Ps30 & pps/psi\\
    NRf & Corrected fan speed & rpm\\
    NRc & Corrected core speed & rpm\\
    BPR & Bypass Ratio & --\\
    farB & Burner fuel-air ratio & --\\
    htBleed & Bleed Enthalpy & -- \\
    Nf\_dmd & Demanded fan speed & rpm \\
    PCNfR\_dmd & Demanded corrected fan speed & rpm\\
    W31 & HPT coolant bleed & lbm/s\\
    W32 & LPT coolant bleed & lbm/s \\
    \toprule
    Environment Parameters & Description & Units \\
    \midrule
    Altitude  & -- & ft.\\
    TRA & Throttle resolver angle &  deg.\\
    Mach number & -- & -- \\
    \bottomrule
\end{tabular}%
  \label{cmapss_dataset_explanation}
\end{table}

The turbofan engine consists of several components, which are illustrated in figure \ref{turbofan_diagram}. All datasets provide data for 21 sensors and three environment parameters, which change with time step. The values of the three environment parameters determine which specific operating condition the engine is in.  The statistics of the C-MAPSS and PHM08 datasets are shown in Table \ref{cmapss_dataset_statistic}. Each sensor and environment parameter explanation is given in Table \ref{cmapss_dataset_explanation}. For each engine, the engine is healthy at the beginning of the simulation. At some random time step, a failure occurs, which propagates, causing the engine to degrade until it is completely damaged. The dataset consists of run-to-failure trajectories for multiple engines. The train set gives complete data of the trajectories, while in the test set, only part of the data is given, and the researcher is required to predict the amount of time steps the engine can still run.\par

\begin{figure}[htbp]
  \centering
  \includegraphics[width=\textwidth]{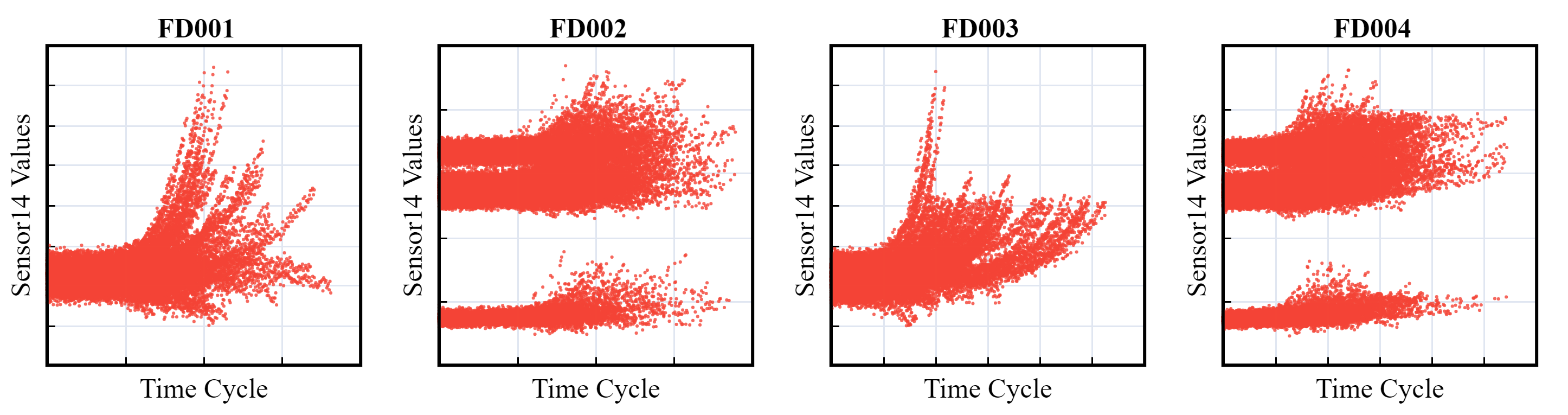}
  \caption{Scatter plot of Sensor 14 data distribution on the C-MAPSS dataset.}
    \label{sensor14_scatter}
\end{figure}
Four sub-datasets of C-MAPSS are four combinations of fault modes and working conditions. Since working conditions significantly impact sensor data, more engines are given for the FD002 and FD004 datasets with six working conditions. Figure \ref{sensor14_scatter} shows the data distribution of the 14th sensor on FD001 -- FD004. It can be found that the engine degradation proceeds in different directions due to the different fault modes and working conditions in the four data sets, which leads to different data distribution in each dataset.\par

\subsection{Data Pre-Processing}
Data pre-processing has a crucial impact on the performance of the RUL prediction model. In the experiments, a time window technique is used to segment the data. Piece-wise RUL modeling is applied to the degradation process, and regularization is used to normalize the data. For datasets with multiple working conditions, a k-means clustering is also performed to environment parameters.\par

\subsubsection{Piece-Wise RUL}
\begin{figure}[htbp]
  \centering
  \includegraphics[scale=0.9]{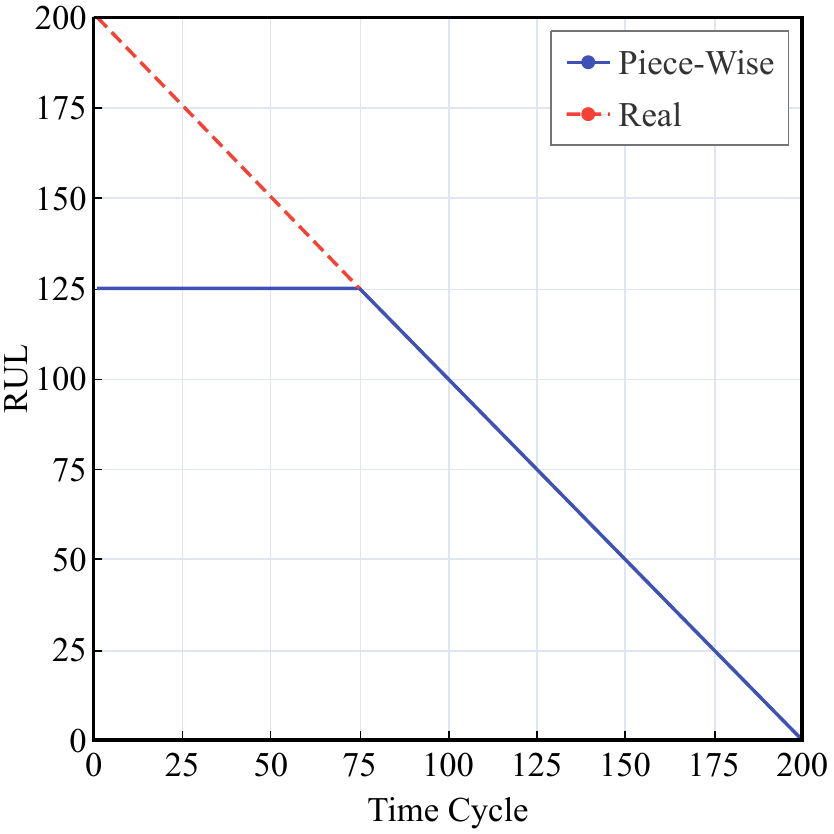}
  \caption{Piece-wise RUL Function.}
    \label{piece_wise_rul}
\end{figure}
Since the engine fails at some unknown time and is healthy before the failure, it is also difficult to predict the exact RUL of a healthy engine. Thus a linear degradation model is essential for the convergence of the model. In piece-wise RUL, shown in Figure \ref{piece_wise_rul}, a constant value $R_{max}$ is set to indicate the remaining useful life of a healthy engine. In the beginning, the RUL of the engine is constant and then begins to degrade linearly. Piece-wise RUL is proved to be effective for the C-MAPSS dataset as it helps the network to converge better\cite{cnn2d}\cite{attn_lstm}. Following other studies, $R_{max}$ is set to 125 in this experiment. We will discuss the effect of the $R_{max}$ setting on the model performance in subsequent experiments.\par

\subsubsection{Cluster and Normalization}
\begin{figure}[htbp]
  \centering
  \includegraphics[width=0.4\textwidth]{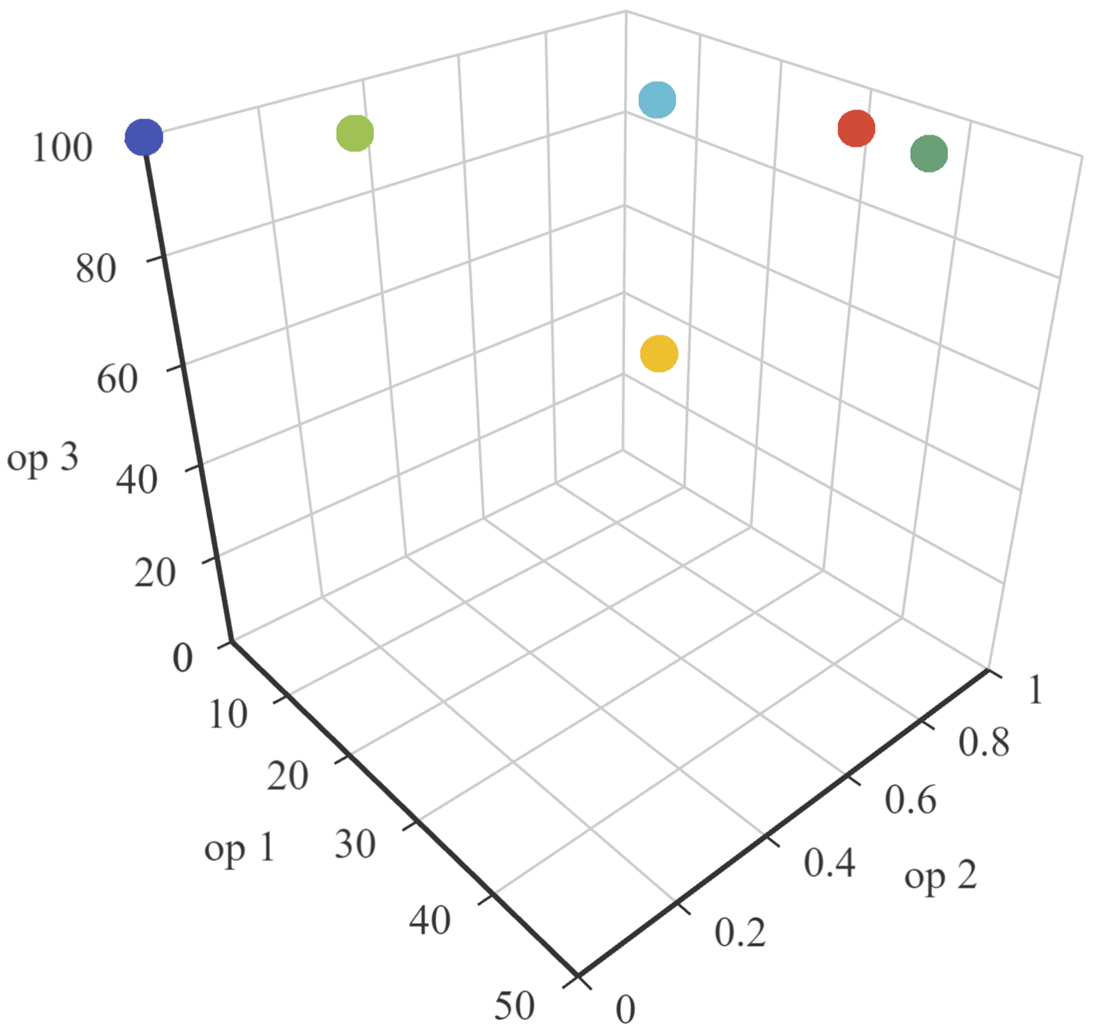}
  \caption{Clustering results of environment parameters on FD004 dataset.}
    \label{cluster_scatter}
\end{figure}
The outputs of different sensors need to be normalized because of the different data units. For C-MAPSS and PHM08, the two most commonly used methods are min-max normalization\cite{cnn2d} and z-score normalization\cite{lstm2}. In the absence of prior expertise, reasonable max-min values for sensors are difficult to determine, so we use z-score normalization, which is defined as follows:
\begin{equation}
\label{z_score_norm_together}
    z_i=\frac{x_i-\mu_i}{\sigma_i}, i\in S
\end{equation}
where $S$ denotes the collection of sensors, and $z_i$ represents the normalized value of the $i$-th sensor, $\mu_i$ and $\sigma_i$ represent the average and standard deviation of the $i$-th sensor, respectively.\par

\begin{figure}[htbp]
  \centering
  \resizebox{\columnwidth}{!}{
  \includegraphics{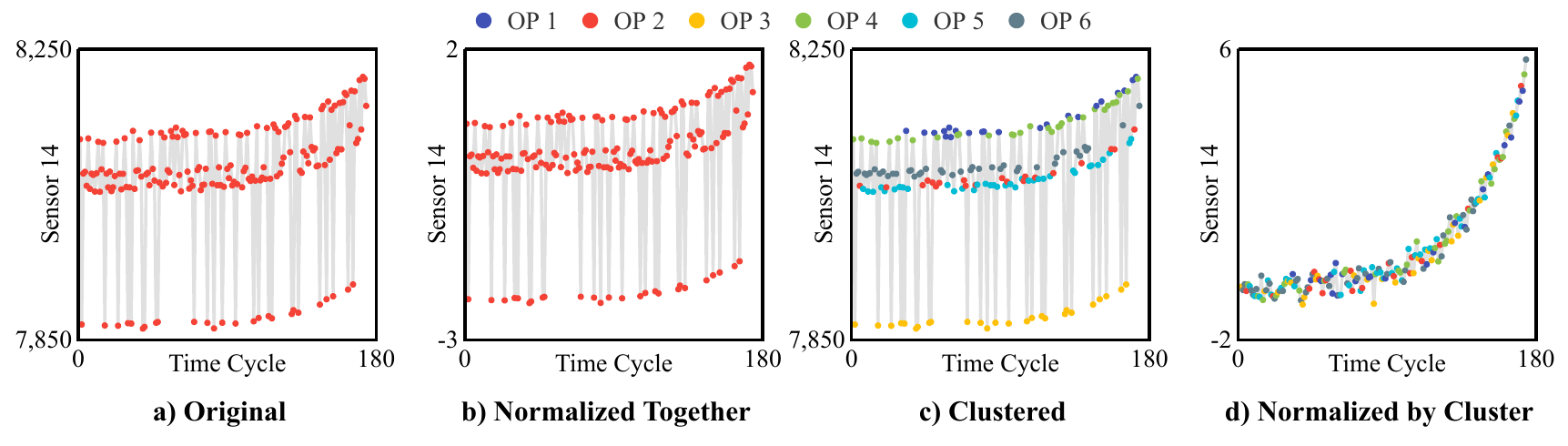}
  }
  \caption{Original and normalized trajectory for the 14th sensor of the No.7 engine in the FD002 dataset.}
  \label{normalize_exaple}
\end{figure}

Various working conditions have a substantial influence on the sensor data. So it is essential to process the sensor data separately according to the different operational settings. Take the 14th sensor of an engine on FD002 as an example. Figure \ref{normalize_exaple}.a) shows the original data of the sensor in degradation progress, where the series data violently oscillates, resulting in the overall trend becoming obscure. Figure \ref{normalize_exaple}.b) is the data normalized by formula \ref{z_score_norm_together}, which changed in the data range but did not change in data distribution. Therefore, we normalize the data separately according to the working conditions at the time of data recording so that the normalized data can reflect the trend of degradation.\par

We can easily get the classification of six conditions by clustering based on the three operational settings given in the datasets. The results of clustering three operational settings using the K-means algorithm are shown in Figure \ref{cluster_scatter}. The values of the different working conditions are very centralized and separated to a large extent, so the clustering results are highly reliable.\par

After getting the working conditions of the data for each time step, we can normalize the data according to the working condition separately as the following equation:

\begin{equation}
    z_{i,j}=\frac{x_{i,j}-\mu_{i,j}}{\sigma_{i,j}}, i\in S, \ j \in C
\end{equation}
where $C$ represents the set of conditions, $i$ represents the $i$-th sensor, and $j$ represents the $j$-th conditions. Figure \ref{normalize_exaple}.c) illustrates the data colored by working conditions, and \ref{normalize_exaple}.d) is the data normalized separately by working conditions, which shows that the data trend with the time step is distinct from the unnormalized data.\par
\subsubsection{Time Window}

\begin{figure*}[htbp]
  \centering
 
  \resizebox{\columnwidth}{!}{
 \includegraphics{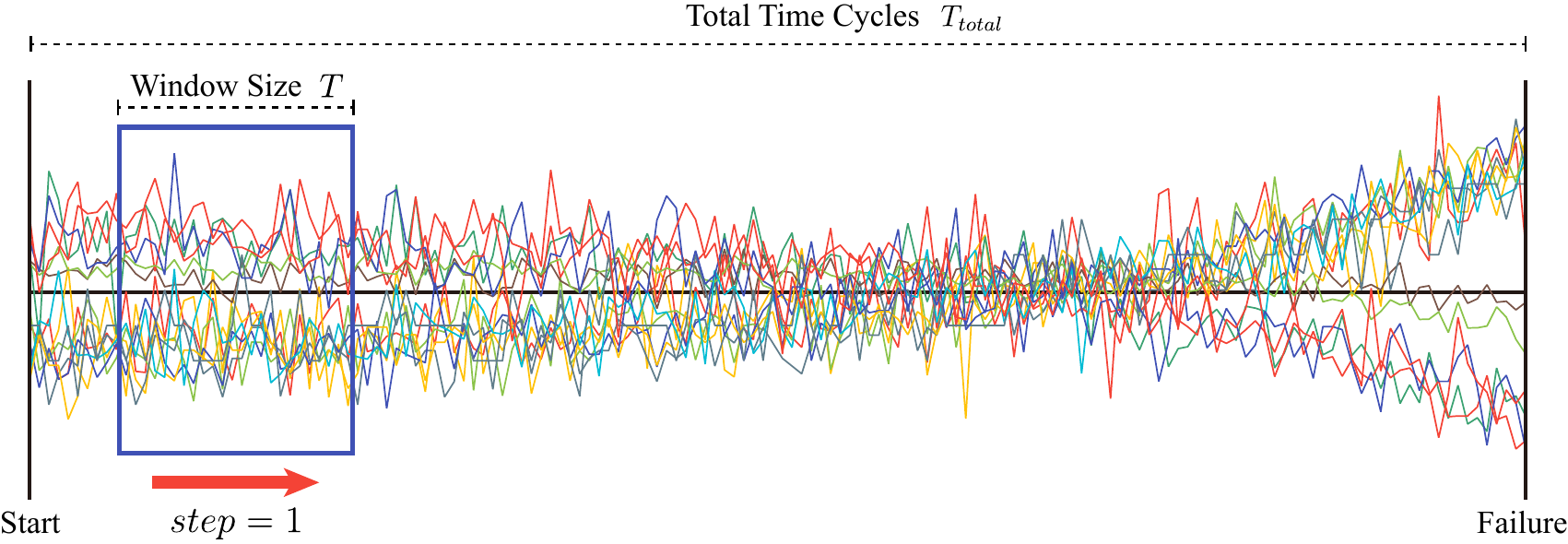}
}
  \caption{Illustration of splitting data with the time window method.}
    \label{time_window_diagram}
\end{figure*}

Time window is a common technique for data segmentation, which is illustrated in Figure \ref{time_window_diagram}. For each run-to-failure time series, a time window slides from the beginning to the end. Let the total sequence length be $T_{total}$, the window length be $T$, and the step size of the window sliding is fixed to 1. Then, the RUL of the engine at $t_i$ can be expressed as
\begin{equation}
    RUL_i=T_{total}-ti \qquad t_i =1,2,3,...,T_{total}
\end{equation}
The number of samples obtained for each sequence can be expressed as
 \begin{equation}
    Sample \ number=\begin{cases}
T_{total} - T + 1 & T \le T_{total}\\
1 & T > T_{total}
\end{cases}
 \end{equation}\par
As the test set does not give the whole sequence, the sequence length T may be smaller than the window length. The number of samples at this point is one, and the sequences are filled forward to the window length using the data of the first time step.\par

\subsection{Setup}
\subsubsection{Metrics}

Two widely adopted metrics were used to evaluate the proposed method's performance: the score function and the root mean square error(RMSE). The score function is the evaluation metric used in the PHM08 challenge dataset\cite{cmapss}, which is formulated as follows:
  \begin{equation}
      \begin{split}
          Score=\begin{cases}
 \displaystyle\sum_{i=1}^N e^{-(\frac{d}{a_1})} -1 & for \ d < 0\\
 \displaystyle\sum_{i=1}^N e^{(\frac{d}{a_2})} -1 & for \ d \ge 0
\end{cases}\\
\begin{aligned} where \ 
d =\widehat{RUL} - RUL,\quad a_1 =13, \quad a_2 =10
\end{aligned}
      \end{split}
  \end{equation}
  
The other metric is the RMSE, which has the following equation.
\begin{equation}
    RMSE=\sqrt{\frac{1}{N}\sum_{i=1}^Nd_i^2}
\end{equation}
\par

\begin{figure}[htbp]
  \centering
  \includegraphics[scale=0.9]{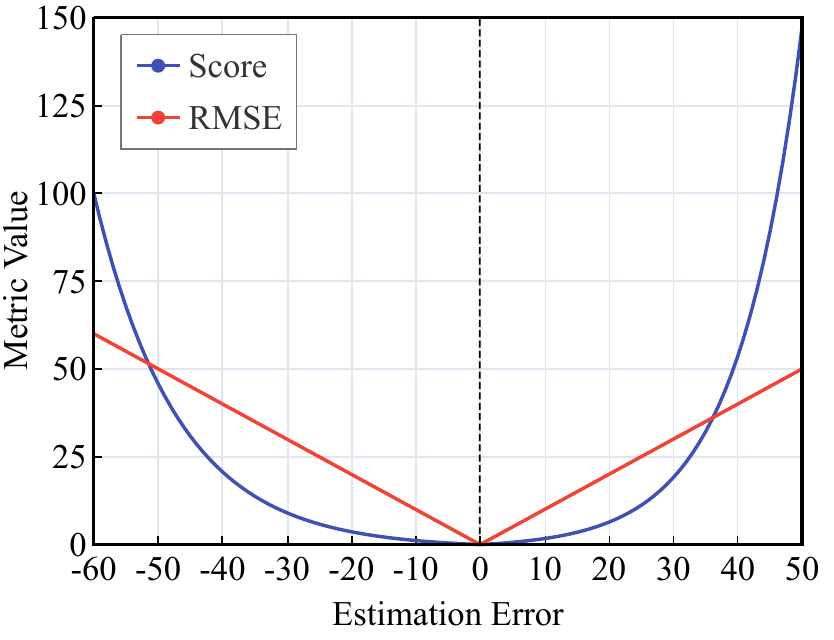}
  \caption{Comparison of RMSE and score functions.}
    \label{rmse_score_compare}
\end{figure}

A comparison of the two metrics is shown in Figure \ref{rmse_score_compare}. With increasing error between the prediction and truth values, the RMSE increases linearly, while the Score score increases exponentially. Moreover, the score function penalizes more severely when the prediction is larger than the real RUL, since predicting a larger RUL for safety-critical industrial equipment usually results in more severe consequences than predicting a smaller RUL.\par

\subsubsection{Hyper-Parameters} \label{section_hyper_param}
Experiments were conducted on all four sub-datasets of the C-MAPSS dataset and PHM08 dataset. The window length was uniformly set to 30, and all 24 features (3ops+21sensors) provided by the dataset were used. The Adam\cite{adam} algorithm was applied to optimize the proposed model. The learning rate of Adam was set to 0.0002. The number of training batches was set to 128, and the early stop mechanism was employed to stop training after 50 rounds without better results.\par
For the proposed model, the LSTM uses three hidden layers with 100 nodes per layer. The MLP layer has a hidden layer with 100 nodes, the activation function is ReLU\cite{relu}, and the dropout is set to 0.5. The final output layer uses a single node to predict the RUL value. We open-source the experiment's code on GitHub\footnote{\url{https://github.com/LazyLZ/multi-head-attention-for-rul-estimation}}, hoping to contribute to the community for better improvement.\par

\section{Results Analysis}
This section provides a complete analysis of the experiment results. All experiments are repeated 30 times independently to ensure the accuracy of the results. \par

\subsection{Parameter Study}

We first investigate the effect of different parameters on the model performance, which include the number of feature heads, the number of sequence heads, the window length, and the piece-wise RUL.\par

\subsubsection{Impact of Feature Head}
\begin{figure*}[htbp]
  \centering
  \resizebox{\columnwidth}{!}{
  \includegraphics{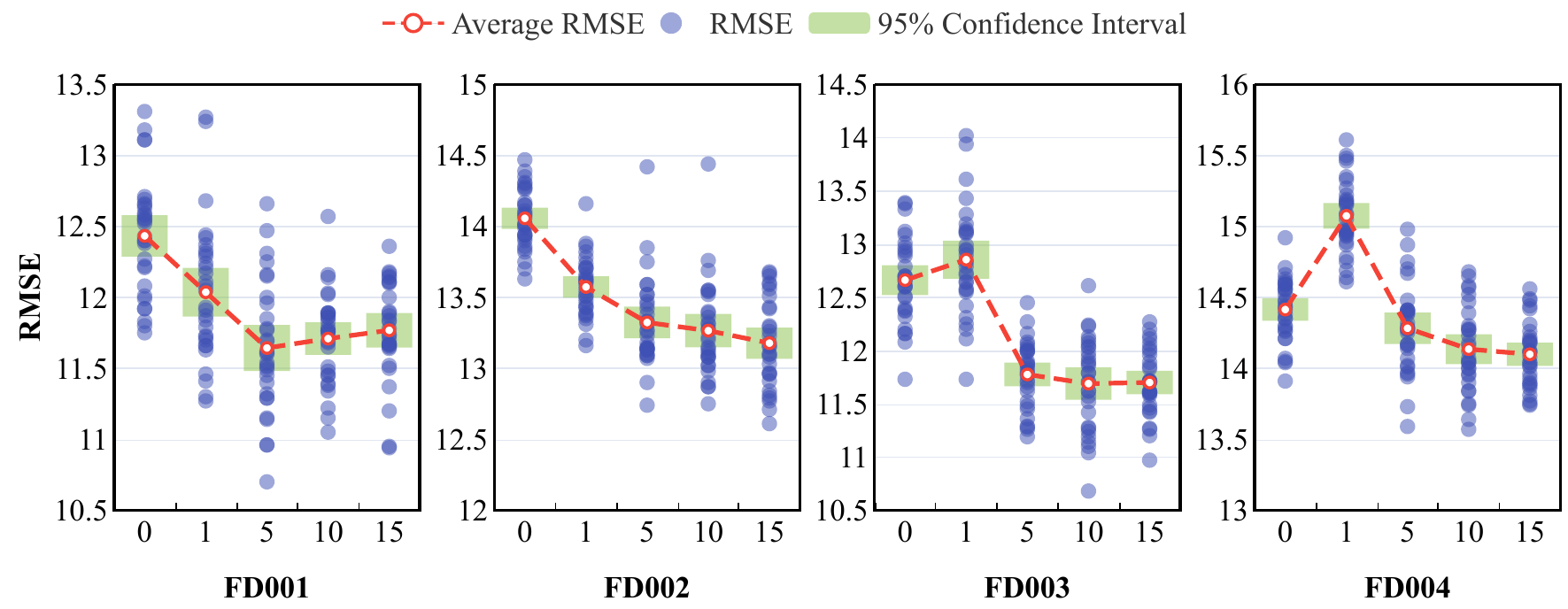}
  }
  \caption{Feature head study on the C-MAPSS dataset.}
    \label{feature_head_study}
\end{figure*}
First, we use a feature attention only with one layer and set the different number of heads to evaluate the impact of feature heads on performance. Note that the head number must be a factor of the embedding dimension. Take the head number as \{0, 1, 5, 10, 15\} where head=0 represents no attention mechanism, i.e., a simple deep LSTM network. After 30 times independent repeated trials, the results are shown in Figure \ref{feature_head_study}. The blue points are the results (RMSE) of each trial, the red points are the average values, and the green area is the 95\% confidence interval of the experimental results. The horizontal coordinate represents the number of feature heads.\par
On the FD001 and FD002 datasets with only one fault mode, the RMSE decreases as the number of heads increases. While on the FD003 and FD004 datasets with two fault modes, the single-head attention's performance is not even as good as the LSTM. However, the model performance is still remarkably improved as the head number increases, indicating that the single-head attention is incapable of handling the degradation trend of multiple fault modes simultaneously, while the multi-head attention is competent. On all four datasets of C-MAPSS, better results are obtained on head=10 or 5, while the gains obtained by further increasing the head count are not significant. Overall, experiments on all four datasets demonstrate that the multi-head attention mechanism outperforms traditional attention mechanisms and significantly contributes to the prediction performance.\par

\subsubsection{Impact of Sequence Head}
\begin{figure*}[htbp]
  \centering
  \resizebox{\columnwidth}{!}{
  \includegraphics{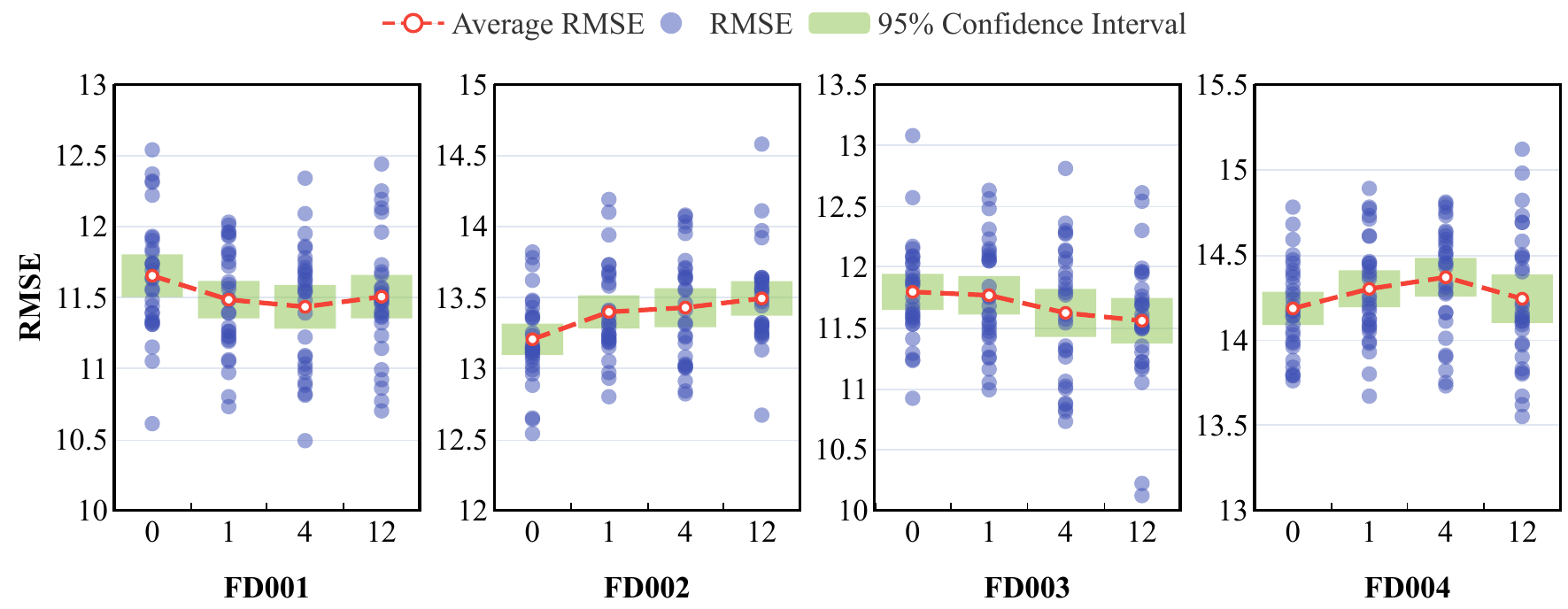}
  }
  \caption{Sequence head study on the C-MAPSS dataset.}
    \label{sequence_head_study}
\end{figure*}
We also study the impact of sequence attention. Based on the study on feature attention, we set the head of feature attention to 5 on FD001 and FD003, 15 on FD002 and FD004, respectively. Then we adjust the sequence attention to \{0, 1, 4, 12\}, where 0 represents only feature attention.\par
The results are shown in Figure \ref{sequence_head_study}. First, the improvement in sequence attention is generally less pronounced than feature attention because the data is generated with noise. For the feature vector consisting of time series, time sequence performs noise reduction, while for sequence attention, the feature vector consisting of data from 24 sensors does not have this effect.\par
On the FD001 dataset, the best results are obtained at head=4. A higher number of heads increases the training time, but with worse results. The results for FD003 are similar to FD001's,  except that the performance has not yet reached the inflection point at head=12. For FD002 and FD004, the performance becomes worse regardless of the number of heads. Therefore, sequence attention has no significant effect on the performance for data with multiple conditions.\par

\subsubsection{Impact of RNN Cell}

\begin{figure*}[htbp]
  \centering
  \resizebox{\columnwidth}{!}{
  \includegraphics{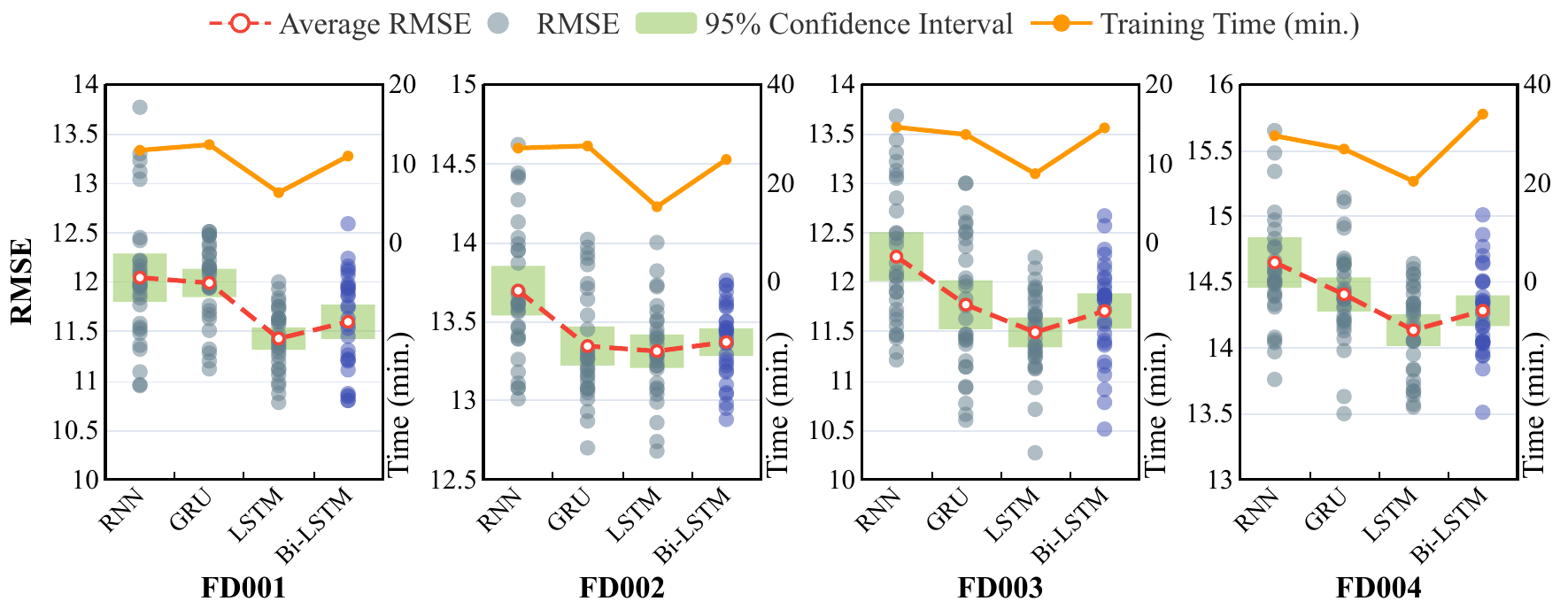}
  }
  \caption{RNN cell study on the C-MAPSS dataset.}
    \label{rnn_cell_study}
\end{figure*}
There are many variants of RNN. In this paper, we select four of the most commonly used structures: basic RNN, GRU, LSTM, and Bidirectional LSTM, to study the effects of different RNN cells on performance. We use the hyper-parameters of the attention layer obtained in the subsections above and only change the cell structure of the RNN. The experiment results are shown in Figure \ref{rnn_cell_study}, where the green intervals, red dots, dashed lines, and blue dots have the same meaning as the experiments above. The orange line represents the average training time of the model for each experiment.\par

Among them, RNN has the worst performance, followed by GRU. Meanwhile, the repeated trials of these two models are not as stable as others. Although the structure of these models is simpler with less computation,  it also leads to difficulties in convergence and more training epochs. Therefore its training time is longer than LSTM and even longer than Bi-LSTM in FD001 and FD002 datasets. Bi-LSTM performs better than GRU but worse than LSTM because the bidirectional LSTM increases the model complexity. Further, the Bi-LSTM is intended to infer the following content through the bidirectional relationship of the sequences. In contrast, the signal sequence of the sensor in the RUL estimation scenario is meaningless when reversed. Therefore BiLSTM did not achieve the desired results.\par

\subsubsection{Impact of Window Size}
\begin{figure*}[htbp]
  \centering
  \resizebox{\columnwidth}{!}{
  \includegraphics{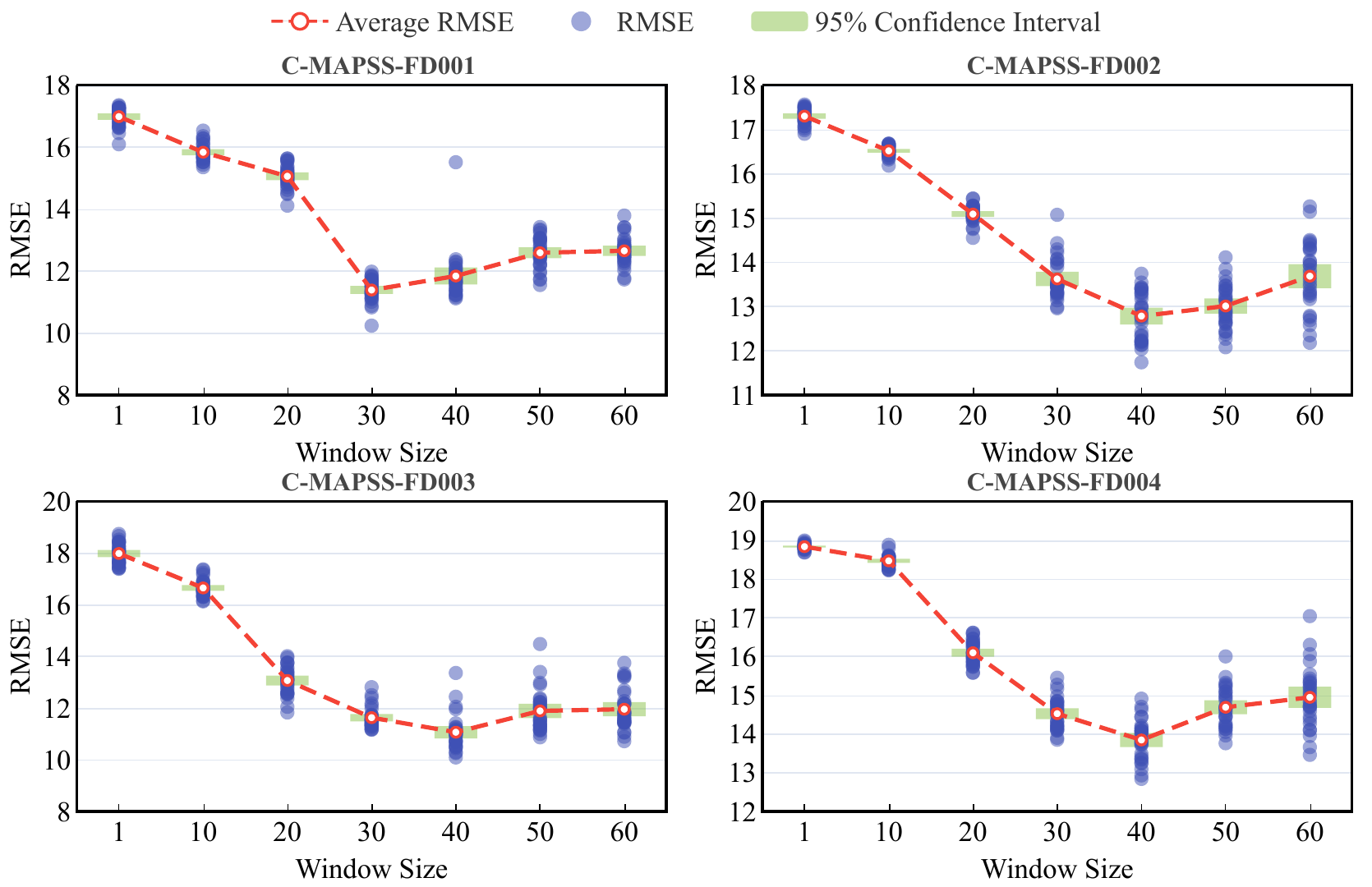}
  }
  \caption{Window size study on the C-MAPSS dataset.}
    \label{window_size_study}
\end{figure*}

Most studies using the C-MAPSS dataset set the window size and $R_{max}$ uniformly to 30 and 125. So we do the same for comparison with other studies. However, we believe that window size and $R_{max}$ are still valuable to investigate.\par
We adjust the window size to \{1, 10, 20, 30, 40, 50, 60\} on four datasets of C-MAPSS, with other parameters as above, and obtain the results in Figure \ref{window_size_study} after 30 independent repeated trials for each configuration. The RMSE first decreases and then increases as the window size increases on all four data sets, which indicates that an excessively long window size does not help the performance. This is because the current health status of the engine is mainly correlated with the data from the most recent period, and the correlation decreases as the time interval becomes larger. Taking these data into consideration may instead introduce too much noise, which is partially evidenced in the experiments. On the other hand, the variance of the trial results gets larger as the window size increases, indicating that the model training is indeed disturbed by noise. On FD001, the best results are achieved with a window size of 30, while on the other three datasets, the best results appear at 40.\par

\subsubsection{Impact of Piece-Wise RUL}
\begin{figure*}[htbp]
  \centering
  \resizebox{\columnwidth}{!}{
  \includegraphics{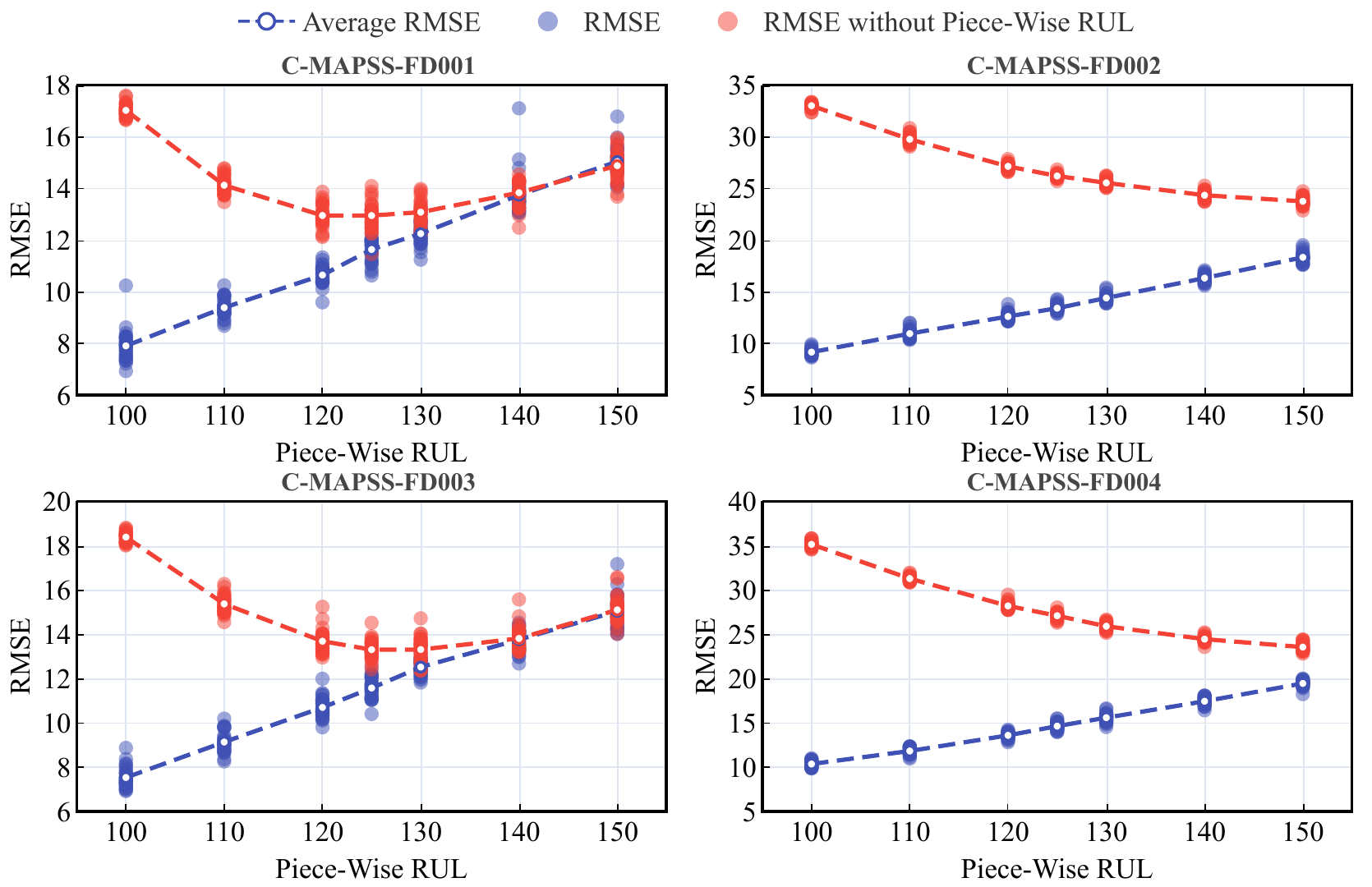}
  }
  \caption{Piece-wise RUL study on the C-MAPSS dataset.}
    \label{piece-wise_RUL_study}
\end{figure*}

Similar to the experiments of window lengths. We set the $R_{max}$ to \{100, 110, 120, 125, 130, 140, 150\}. Each experiment is set to use or not use the piece-wise RUL for the test set separately as a comparison. The results are shown in Figure \ref{piece-wise_RUL_study}, where the red dots and lines are for the test set without piece-wise RUL and blue is for using it.\par
On FD001 and FD003, the results without piece-wise RUL decreases and then rises as $R_{max}$ goes from small to large, while on FD002 and FD004, the RMSE of the red line decreases and the blue line increase as the $R_{max}$ rises. This suggests that a smaller numerical RMSE obtained by setting a very small $R_{max}$ does not indicate that the model has good predictive ability. Because this is achieved by setting too many engines' RUL to $R_{max}$ on the test set. As an extreme example, if $R_{max}$ is set to 0, the test set results will all be 0, and the model will be trained as a model with a constant output of 0, which gives a very low RMSE but does not demonstrate the prediction ability of the model. On the FD001 and FD003 datasets, the best results are indeed obtained when $R_{max}$=125, while on the other two datasets, setting $R_{max}$ to greater than or equal to 150 may be a better choice, although this may not look good numerically.\par

\subsection{Ablation Study}
\begin{figure*}[htbp]
  \centering
  \resizebox{\columnwidth}{!}{
  \includegraphics{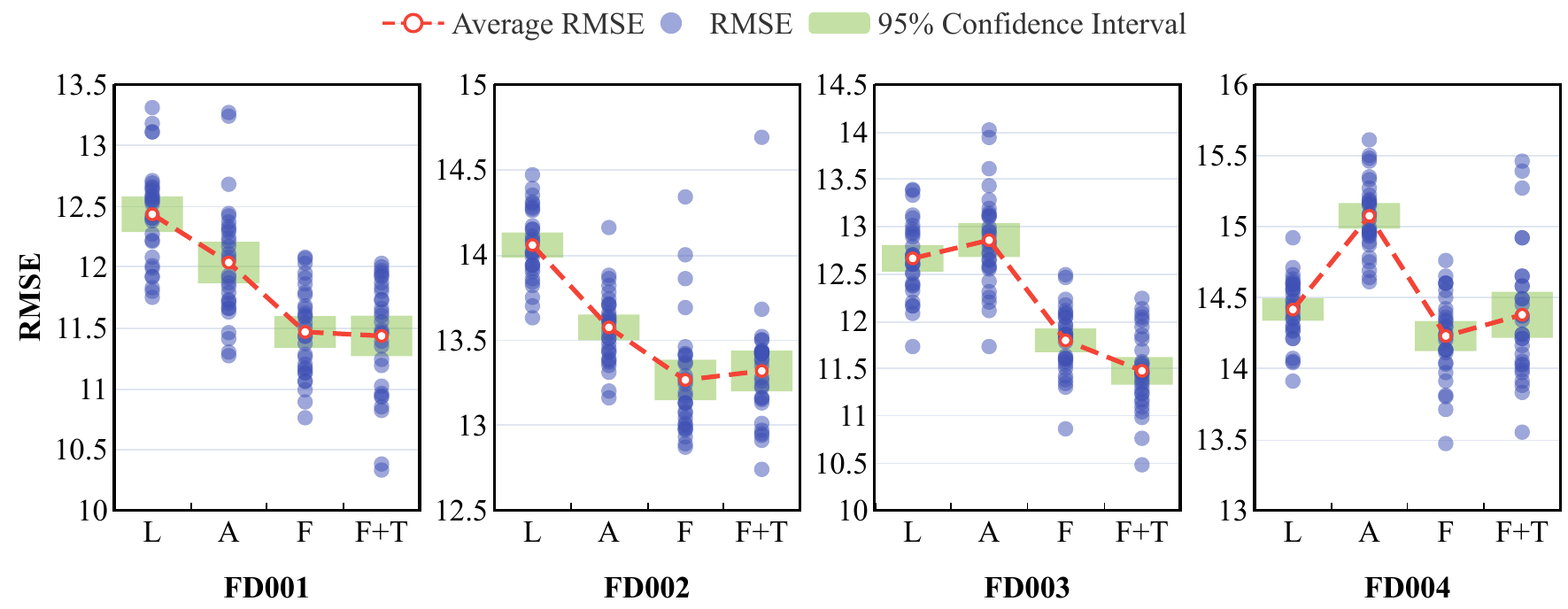}
  }
  \caption{Ablation study on the C-MAPSS dataset.}
    \label{ablation_study}
\end{figure*}
In this section, we design an ablation experiment to verify the effectiveness of each module of the proposed model. The experimental results on the four sub-datasets of C-MAPSS are shown in Figure \ref{ablation_study}. Where $L$ denotes the single LSTM network, $A$ denotes the single head attention mechanism on features, $F$ denotes using only multi-head attention on features, and $F+T$ denotes using multi-head attention both on features and sequences.\par

From the perspective of the fault mode, LSTM performs the worst on FD001 and FD002 with only one fault mode. Single-head attention has a significant performance improvement than LSTM, and multi-head attention mechanism has another enormous improvement compared to single-head attention. As for FD003 and FD004 with multiple fault modes, the single-head attention mechanism is even worse than the simple LSTM, but the model with multi-head attention on feature and the combined model of the two types of attention outperform both the LSTM and the single-head attention model. In terms of working conditions, the model with two types of attention performs best for FD001 and FD003 with only one working condition. On FD002 and FD004 with multiple working conditions, the model with two types of attention is slightly worse than the model with multi-head attention on features.\par

From the above observations, we can conclude that the results of multi-head attention-based models are better than the single-head attention-based model on all four datasets of C-MAPSS since the multi-head attention mechanism learns the weights of the features from different perspectives by using a linear transformation on the inputs. In contrast, single-head attention mechanism cannot consider multiple degradation trends, therefore, the results are not good on the datasets with multiple failure modes (FD003, FD004). The models with two types of attention mechanisms perform poorly on the dataset with multiple working conditions caused by the changes in working conditions that make the sequence data oscillate, resulting in excessive noise learned by the attention mechanism on the sequence.\par

\subsection{Attention Scheme Study}
\begin{figure*}[htbp]
  \centering
  \resizebox{\columnwidth}{!}{
  \includegraphics{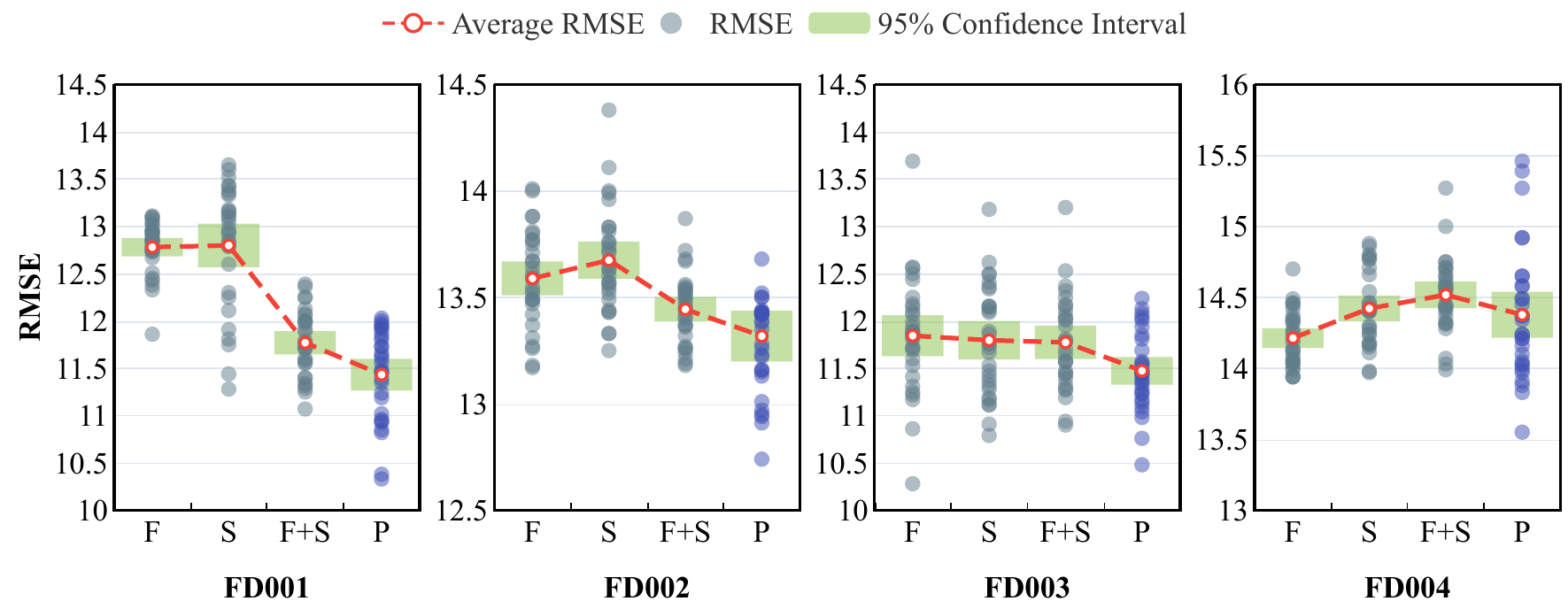}
  }
  \caption{Attention scheme study on the C-MAPSS dataset.}
  \label{other_attention_study}
\end{figure*}
In order to investigate the performance of our model's attention scheme compared to other attention schemes, we referred to the attention mechanism used in \cite{tcn}, weighting the data in different ways and compared it with our model. The experiment results are shown in Figure \ref{other_attention_study}. Where P is the proposed model; F and S are basic attention weighted in feature and time dimension, respectively; F+S represents first weighted by features and then weighted by time. The results show that the basic attention mechanism weighted on features achieves the best results on FD004, probably due to the simple model convergence better. On the other three datasets, the models weighted for both dimensions (F+S, P) are better than those weighted for a single dimension, indicating that attention usage in multiple dimensions contributes to the model performance. Furthermore, our model obtained better results than the model using the basic attention, which demonstrates that the multi-head self-attention performs better than that of the basic attention model.\par

\subsection{Case Study}
\begin{figure*}[htbp]
  \centering
    \resizebox{\columnwidth}{!}{
    \includegraphics{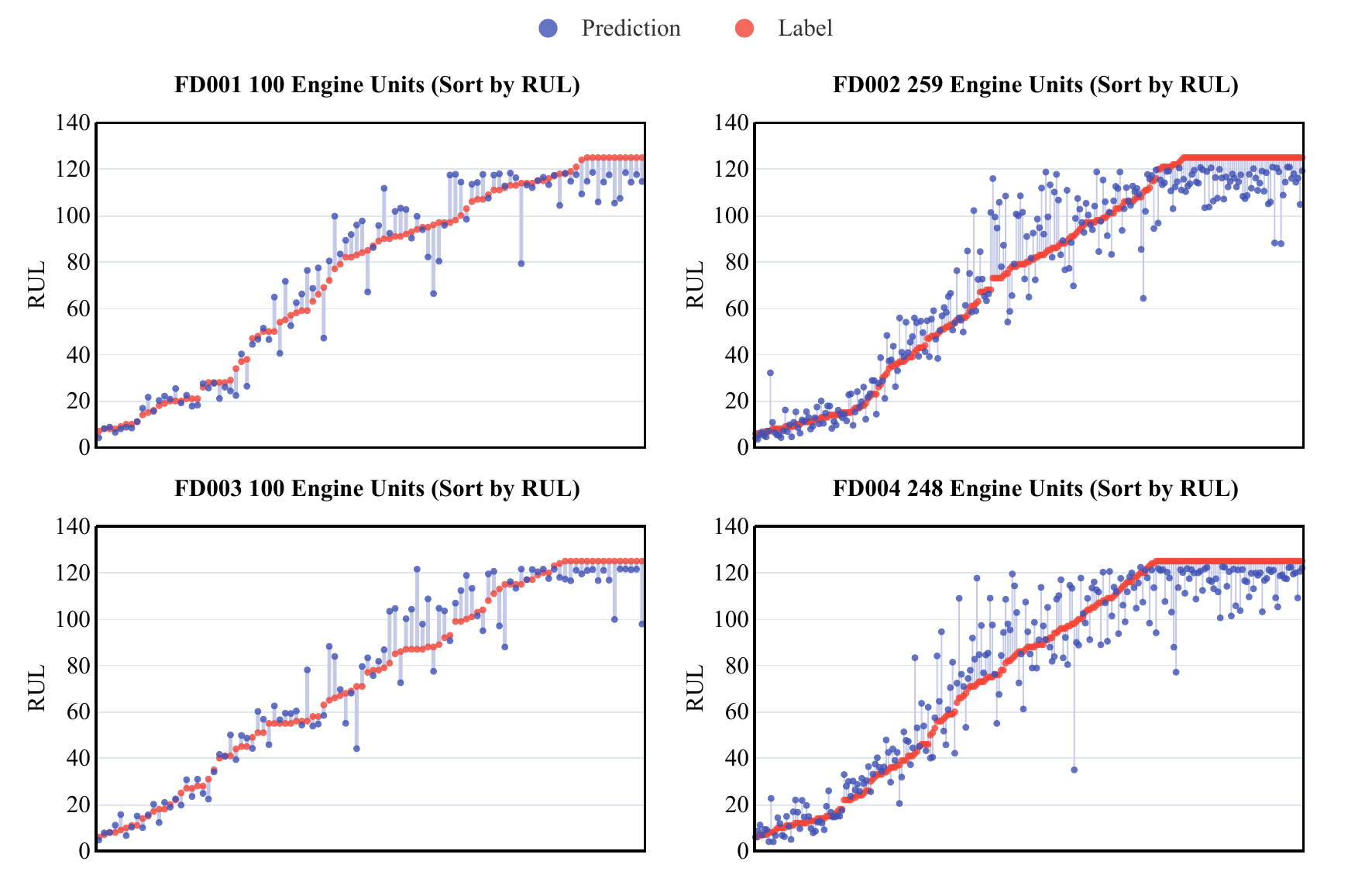}
    }
  \caption{Predictions for all engines on the four datasets.}
    \label{all_engine_result_fd001_fd004}
\end{figure*}
\begin{figure*}[htbp]
  \centering
  \resizebox{\columnwidth}{!}{
  \includegraphics{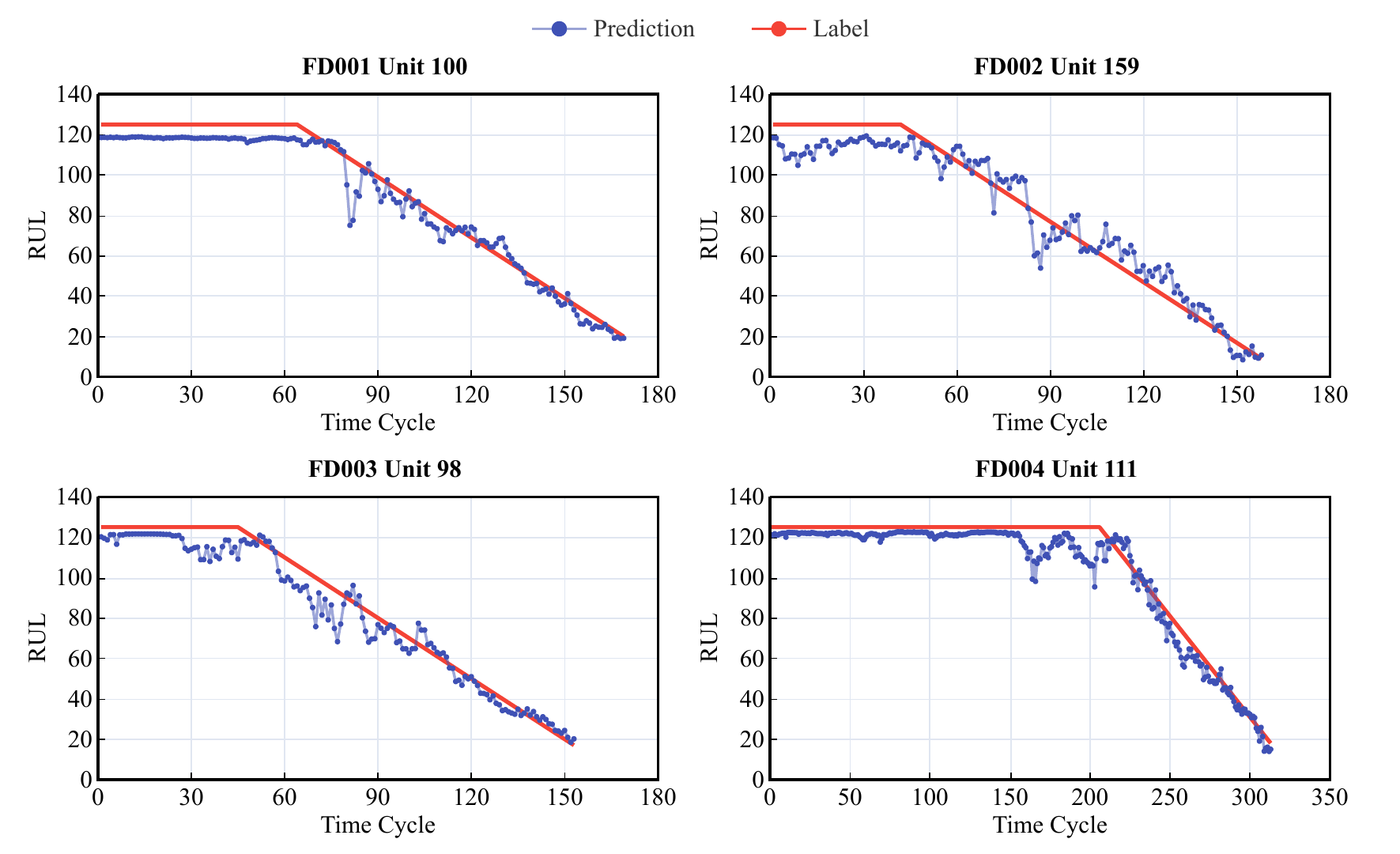}
  }
  \caption{Example of continuous prediction for a single engine.}
  \label{engine_predict_example}
\end{figure*}
Figure \ref{all_engine_result_fd001_fd004} shows the prediction results of all engines on the FD001 -- FD004 test sets. The engines are sorted by real RUL ascendingly in order to observe the model's performance among different RULs better. The model's predictions are generally accurate when RUL is greater than 120 and less than 50 but are comparatively inaccurate between 50 and 120. In the case of a damaged engine, the data of each sensor converge more as the fault spreads, which helps the model predict the RUL more accurately. When the RUL is greater than 120, the model considers the engine to be healthy because of the piece-wise RUL mechanism, which predicts the model's RUL to be $R_{max}$.\par

Although the test set only gives partial sequences, we can calculate the RUL for each time step of the given sequences based on the final RUL. Figure \ref{engine_predict_example} shows the results of all predictions for the four engines selected from FD001\textasciitilde FD004. The proposed model results are stable when continuously predicting the RUL of a standalone engine. As the engine approaches failure, the prediction becomes more accurate, proving the effectiveness of our model in estimating RUL. \par

\subsection{Interpretability of Attention}
The weights of the attention mechanism are interpretable because it uses the softmax function to output probabilities so that we can get some interpretations of the model predictions by the weights of attention. \par
\begin{figure}[htbp]
  \centering
    \includegraphics{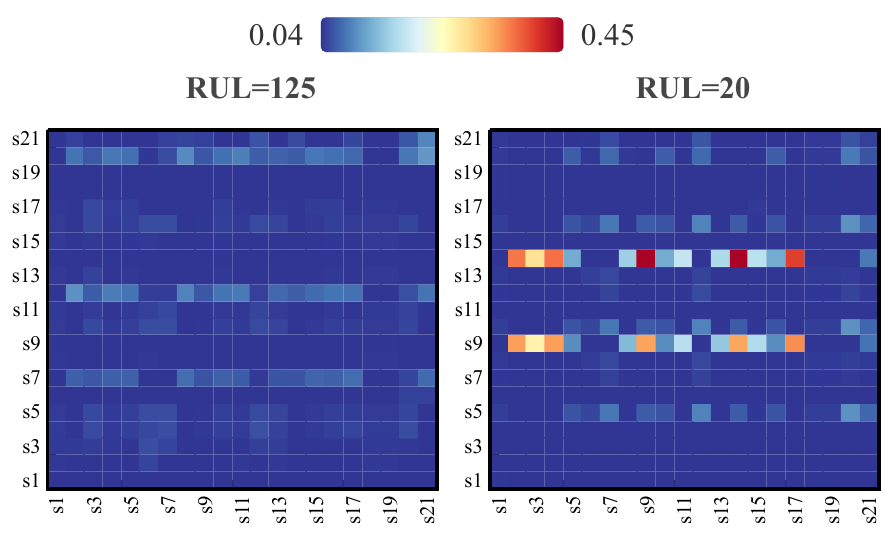}
  \caption{Heat map of the self-attention weight matrix for the No.100 engine of the FD001 test set}
    \label{attention_heatmap}
\end{figure}

Figure \ref{attention_heatmap} shows the heat map of the weights of the self-attention mechanism for the 21 features, where each column represents the corresponding sensor's attention to each sensor. The attention weights are assigned differently for different sensors, which indicates that the self-attention mechanism learns the relationship between sensors. Moreover, at the early stage of degradation, the attention among sensors is relatively even, while the attention of specific sensors increases sharply at the end of degradation. Also, these sensors are commonly attended by other sensors, such as s9, s14 in Figure \ref{attention_heatmap}, indicating that these sensors are vital to the degradation of the engine, and these sensors with higher mutual attention weights may also be correlated in their physical structure.\par
It is noted that the most degraded component is not always the component from which the fault starts. As faults propagate, they always spread to multiple components, and some components are more sensitive to the engine's condition so that they also receive more attention weight.\par
\begin{figure}[htbp]
  \centering
    \includegraphics{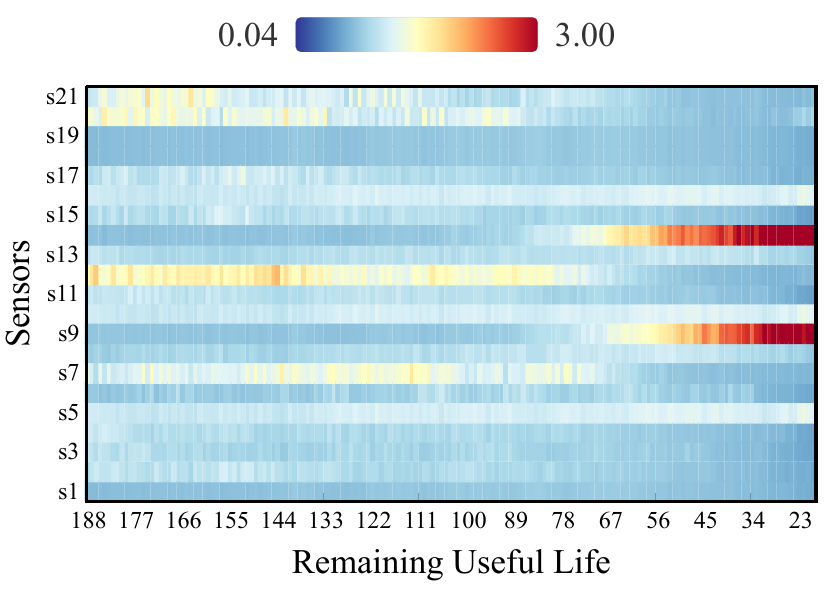}
  \caption{Heat map of the weight sum for the No.100 engine of the FD001 test set}
    \label{attention_cycle_heatmap}
\end{figure}

For each time step, by summing the attention of each sensor, we can obtain the trend of the attention of the sensors over the time step. As shown in Figure \ref{attention_cycle_heatmap}, the attention is distributed relatively evenly at the beginning. While at the end of the degradation, the attention shifts rapidly to several sensors. This may indicate the propagation process of degradation, and the sensors with attention may be more indicative of the increased effect on RUL, i.e., more severe degradation. Moreover, we can find that the shift in attention always occurs in the last 30-40 time steps, which supports the reason why the model achieves the best results when the window size is taken as 30-40 because the data from these last time steps are genuinely representative of the current engine state.\par
\begin{figure}[htbp]
  \centering
    \includegraphics{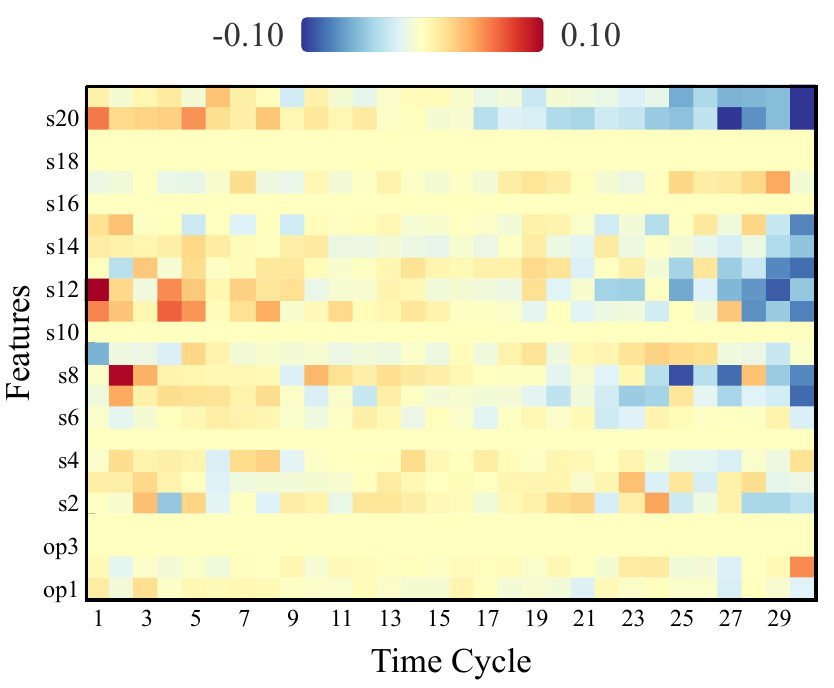}
  \caption{Heat map using SHAP analysis on FD001 test set}
    \label{shap_result}
\end{figure}

Finally, we use SHAP\cite{shap}, a general interpretable method for machine learning, to explain our model. The model and experiment results of the FD001 test are used to analyze the contribution of each feature of the input to the prediction results, shown in figure \ref{shap_result}. The x-axis and y-axis denote time steps and features, respectively. Each square of the heat map is the model input at the corresponding position. The bluer color of the box means that the input at that position is negatively correlated with the results, and contrarily, the redder the positive correlation. As illustrated by the heat map, the most concerned data of the model are the tails and heads of the input sequence, with the tail of the sequence being particularly essential, which is substantially consistent with the fact. From the sensor's perspective, the results of the SHAP analysis also corroborate the effectiveness of our attention mechanism. The attention mechanism concerns sensors such as S9 and S14, which are also those inputs in SHAP analysis that significantly impact the experiment results.\par

\subsection{Comparison with other work}

\subsubsection{Results on C-MAPSS}
\begin{table*}[b!]
  \setlength{\tabcolsep}{12pt}
  \renewcommand{\arraystretch}{1.2}
  \centering
  \caption{Results comparison of different methods on C-MAPSS datasets. }
  \resizebox{\columnwidth}{!}{
  \begin{tabular}{lccccccccc}
    \toprule
    \multicolumn{1}{c}{\multirow{2}[1]{*}{Methods}} & \multirow{2}[1]{*}{Year} &
    \multicolumn{2}{c}{FD001} & \multicolumn{2}{c}{FD002} & \multicolumn{2}{c}{FD003} & \multicolumn{2}{c}{FD004} \\
    \cmidrule{3-10}
          &       & RMSE  & Score & RMSE  & Score & RMSE  & Score & RMSE  & Score \\
    \midrule
    LSTM\cite{lstm1}  & 2017  & 16.14 & 338   & 24.49 & 4450  & 16.18 & 852   & 28.17 & 5550 \\
    DCNN\cite{cnn2d}  & 2017  & 12.61 & 273   & 22.36 & 10412 & 12.64 & 284   & 23.31 & 12466 \\
    Bi-LSTM\cite{bilstm} & 2018  & 13.65 & 295   & 23.18 & 4130  & 13.74 & 317   & 24.86 & 5430 \\
    DAG\cite{dag}   & 2019  & 11.96 & 229   & 20.34 & 2730  & 12.46 & 284   & 22.43 & 3370 \\
    HDNN\cite{hdnn}  & 2019  & 13.02 & 245   & \underline{15.24} & \underline{1282}  & 12.22 & 287   & \underline{18.16} & \underline{1527} \\
    GNMR\cite{gnmr}  & 2020  & 12.14 & \textbf{212}   & 20.85 & 3196  & 13.23 & 370   & 21.34 & 2795 \\
    Attn-LSTM\cite{attn_lstm} & 2020  & 14.53 & 322   & -     & -     & -     & -     & 27.08 & 5649 \\
    AGCNN\cite{attn_cnn_gru} & 2020  & 12.42 & 226   & 19.43 & 1492  & 13.39 & \underline{227}   & 21.5  & 3392 \\
    DA-TCN\cite{tcn} & 2020  & \underline{11.78} & 229   & 16.95 & 1842  & \textbf{11.56} & 257   & 18.23 & 2317 \\
    \textbf{Proposed (F)} & 2021  & \textbf{11.71} & \underline{223}   & \textbf{13.26} & \textbf{1077}   & \underline{11.69} & \textbf{192}   & \textbf{14.14} & \textbf{1375} \\
    \textbf{Proposed (F+T)} & 2021  & \textbf{11.43} & \underline{209}   & \textbf{13.32} & \textbf{1058}   & \underline{11.47} & \textbf{187}   & \textbf{14.38} & \textbf{1618} \\
    \bottomrule
\end{tabular}%
  }
  \label{table_cmapss_result_comparison}%
\end{table*}%
Since studies of RUL prediction widely use the C-MAPSS dataset, we can make a side-by-side comparison of these works. The comprehensive comparison of the proposed method with state-of-art studies is presented in Table \ref{table_cmapss_result_comparison}. Bold font indicates the best result, and underlined indicates the second-best result. The proposed method achieves the best results in most cases. The differences in the results obtained on the four datasets are due to the different fault modes and working conditions. First, FD002 and FD004 have significantly worse metrics than FD001 and FD003 in all studies because the variation in working conditions causes the serial data to lose its original trajectory with component degradation. Even by regularizing the data, the oscillations due to changes in operating conditions cannot be completely neutralized.

In all studies, the results for FD001 and FD003 are similar, which indicates that the fault modes have little effect on performance. Thus the model should achieve comparable results for a wide range of fault modes that may occur in practice. Our study achieves outstanding results for FD002 and FD004 due to normalization based on the working conditions separately so that the degradation trend that would have been disturbed by the working conditions can be recovered to some extent, which makes the model can learn the degradation trend of the sequence better.\par

\subsubsection{Results on PHM08}
\begin{table}[htbp]
  \setlength{\tabcolsep}{12pt}
  \renewcommand{\arraystretch}{1.2}
  \centering
  \caption{Results comparison of different methods on PHM08 dataset. }
  \begin{tabular}{ l c }
\toprule
 Methods & Score\\ 
\midrule
 SVR\cite{cnn1d} & 15886 \\  
 RVR\cite{cnn1d} & 8242 \\
 MLP\cite{cnn1d} & 3212 \\  
 CNN\cite{cnn1d} & 2056 \\
 LSTM\cite{lstm1} & 1862 \\  
 Attn-LSTM\cite{attn_lstm} & \underline{1584} \\
 Proppsed & \textbf{1060} \\
\bottomrule
\end{tabular}
  \label{table_phm08_result_comparison}%
\end{table}%
We also validate our model on the PHM08 dataset. The test set results for the PHM08 dataset are not given directly but need to be submitted to the official website to receive the score values. The data for the PHM08 challenge dataset was generated from on fault mode and six working conditions, which is consistent with the FD002 dataset. Note that the test set used for the competition had 435 engines, while the current website provides a dataset of 218 engines for the evaluation. Table \ref{table_phm08_result_comparison} shows the comparison of our model with other methods. Compared with CNN, LSTM, attention-based LSTM, and other methods, our model achieves better result, which is consistent with our experimental results on the C-MAPSS dataset.\par

\section{Conclution}
This paper proposes a data-driven deep learning model for predicting the remaining useful life of complex machines containing multiple sensors. The proposed model learns data features through a multi-dimensional self-attention mechanism. In particular, attention on feature dimension is used to learn interactions between model features, and attention on sequences is used to learn the influence weights at different time steps. An LSTM network is applied to learn the sequential features. Finally, an MLP is used to get the final RUL results.\par

Extensive experiments on the C-MAPSS dataset and PHM08 dataset demonstrate the effectiveness of our model in dealing with the RUL estimation tasks for multi-dimensional time series. Time windows, piece-wise RUL, z-score regularization, and K-mean clustering are utilized to pre-process the data. On FD002 and FD004 datasets, normalizing data by working conditions respectively make a vast improvement to performance.

Unlike \cite{attn_lstm}\cite{cnn2d}\cite{tcn}, the proposed method does not require feature selection which makes it more generalizable. We investigate the impact of window size, the number of heads in the multi-head attention mechanism, and $R_{max}$ on the model and use them to tune our model performance to the optimum. We also demonstrate that each part of our proposed model contributes to the performance through the ablation experiments. Preliminary studies on the interpretability of the attention mechanism show that the multi-head attention of the proposed model is able to focus on the degradation trend of the engine, which is very promising for interpretable prognosis and health maintenance systems.\par

In our future work, we will continue focusing on the application of the attention mechanism in RUL estimation, such as Transformer and its variant structures. Graph neural networks are also one of the rapidly developing branches of deep learning, and we will later try GNN-based model structures as well. Finally, we will try to apply our models to more industrial scenarios.\par









\section*{Acknowledgment}
This work was supported in part by the National Natural Science Foundation of China under Grant 61202445 and 2018A030313354, and the CERNET Innovation Project under Grant NGII20180404.




\bibliographystyle{elsarticle-num} 
\bibliography{references.bib}

\end{document}